\documentclass[journal]{IEEEtran}
\usepackage[space]{grffile}
\usepackage{graphicx, amssymb, times}  
\usepackage{amsmath}
\usepackage{algorithm}
\usepackage{algpseudocode}
\usepackage{microtype}
\usepackage{hyperref}
\usepackage{scalerel}
\usepackage{tikz}
\usetikzlibrary{svg.path}

\definecolor{orcidlogocol}{HTML}{A6CE39}
\tikzset{
  orcidlogo/.pic={
    \fill[orcidlogocol] svg{M256,128c0,70.7-57.3,128-128,128C57.3,256,0,198.7,0,128C0,57.3,57.3,0,128,0C198.7,0,256,57.3,256,128z};
    \fill[white] svg{M86.3,186.2H70.9V79.1h15.4v48.4V186.2z}
                 svg{M108.9,79.1h41.6c39.6,0,57,28.3,57,53.6c0,27.5-21.5,53.6-56.8,53.6h-41.8V79.1z M124.3,172.4h24.5c34.9,0,42.9-26.5,42.9-39.7c0-21.5-13.7-39.7-43.7-39.7h-23.7V172.4z}
                 svg{M88.7,56.8c0,5.5-4.5,10.1-10.1,10.1c-5.6,0-10.1-4.6-10.1-10.1c0-5.6,4.5-10.1,10.1-10.1C84.2,46.7,88.7,51.3,88.7,56.8z};
  }
}

\newcommand\orcidicon[1]{\href{https://orcid.org/#1}{\mbox{\scalerel*{
\begin{tikzpicture}[yscale=-1,transform shape]
\pic{orcidlogo};
\end{tikzpicture}
}{|}}}}

\usepackage{cite}

\ifCLASSOPTIONcompsoc
  \usepackage[caption=false,font=smaller,labelfont=sf,textfont=sf]{subfig}
\else
  \usepackage[caption=false,font=scriptsize]{subfig}
\fi

\usepackage{url}

\begin{document}

\title{{\em e-TLD}: Event-based Framework for Dynamic Object Tracking}

\author{Bharath~Ramesh~\orcidicon{0000-0001-8230-3803},~\IEEEmembership{Member,~IEEE},
        Shihao~Zhang~\orcidicon{0000-0003-2597-0444},
				Hong~Yang~\orcidicon{0000-0002-8879-5025},
			Andres~Ussa~\orcidicon{0000-0001-8112-6681},
				Matthew~Ong~\orcidicon{0000-0002-7865-6355}, Garrick~Orchard~\orcidicon{0000-0002-1243-2711},
        and~Cheng~Xiang~\orcidicon{0000-0002-1229-6860},~\IEEEmembership{Member,~IEEE}
\thanks{B. Ramesh is with the N.1 Institute for Health (formerly Singapore Institute of Neurotechnology), National University of Singapore, Singapore 117456. E-mail: bharath.ramesh03@u.nus.edu.}
\thanks{A. Ussa is with the N.1 Institute for Health. H. Yang and G. Orchard were with the Temasek Laboratories, National University of Singapore, when this work was completed. S. Zhang, M. Ong and C. Xiang were with the Department of Electrical and Computer Engineering, National University of Singapore, when this work was done. }
\thanks{Manuscript received.}}

\markboth{IEEE Trans. on Circuits and Systems for Video Technology}%
{Ramesh \MakeLowercase{\textit{et al.}}: {\em e-TLD}: Event-based Framework for Dynamic Object Tracking}
%

\maketitle

\begin{abstract}
This paper presents a long-term object tracking framework with a moving event camera under general tracking conditions. A first of its kind for these revolutionary cameras, the tracking framework uses a discriminative representation for the object with online learning, and detects and re-tracks the object when it comes back into the field-of-view. One of the key novelties is the use of an event-based {\em local sliding window} technique that tracks reliably in scenes with cluttered and textured background. In addition, Bayesian bootstrapping is used to assist real-time processing and boost the discriminative power of the object representation. On the other hand, when the object re-enters the field-of-view of the camera, a \emph{data-driven, global sliding window} detector locates the object for subsequent tracking. Extensive experiments demonstrate the ability of the proposed framework to track and detect arbitrary objects of various shapes and sizes, including dynamic objects such as a human. This is a significant improvement compared to earlier works that simply track objects as long as they are visible under simpler background settings. Using the ground truth locations for five different objects under three motion settings, namely translation, rotation and 6-DOF, quantitative measurement is reported for the event-based tracking framework with critical insights on various performance issues. Finally, real-time implementation in C++ highlights tracking ability under scale, rotation, view-point and occlusion scenarios in a lab setting. 
\end{abstract}


\IEEEpeerreviewmaketitle

\section{Introduction}
\label{sec:intro}
\IEEEPARstart{S}{tandard} video cameras struggle to capture crisp images of scenes characterized by high dynamic range and motion, returning blurred or saturated images. To overcome these limitations, event cameras aim to emulate the important asynchronous property of the human retina, thus earning themselves the name ``silicon retinas''. Hence, an event camera has no global clock or shutter to record images in the traditional sense. Instead, each pixel individually adapts and responds to temporal changes in log intensity, and outputs an asynchronous event with the pixel address which gets a precise timestamp in the order of microseconds. 
\par
An event is characterized by a spatial location ($x,y$), timestamp $t$ and a binary-valued polarity $p$, i.e., on-events ($p=1$) are caused by a positive change in log-intensity and vice-versa for off-events ($p=0$). In both cases, events triggered by brightness changes are likely to occur at the edges that delineate the structure of the scene, and thus removing redundancy with a much lower data rate compared to a standard VGA resolution video at 30 fps. Although redundancy is absent in the event stream, the higher time-resolution should in principle contain all the information of a standard video without bounds on frame-rate and dynamic range. The image reconstruction from a pure event stream lends support to this idea \cite{Kim2014}. 
\par
Despite object tracking being a major research topic in computer vision, applicability has been limited by the low camera dynamics of standard vision sensors. Increasing the frame rate only burdens the computation techniques \cite{Watanabe2014,SMC1,SMC2}, preventing a dynamic, low-latency formulation of object tracking. On the other hand, the discontinuous motion information captured using a standard 30 fps video camera is an obvious disadvantage for frame-based object tracking algorithms. 
\par
This paper introduces a simple and efficient object tracking framework, consisting of a local \emph{tracker} and a global \emph{detector}, by taking advantage of the sparsity and higher temporal resolution of the event camera. In other words, the position of the object in the field-of-view of the event camera changes with negligible spatio-temporal discontinuity (5-10 $\mu$s). Therefore, the key idea is to spatially limit the search region of the tracked object while the temporal limits are imposed by the rate at which events arrive within the search region. In particular, the tracker search is modeled as a discriminative classification scheme (object vs. background) using the event-based descriptor proposed in \cite{Ramesh2017a}. Therefore, given the initial location of the object within a short time-interval, the training phase of the tracker learns a binary classifier. Subsequently, an object detector is learned using the training samples of the tracker. 
\par
The proposed framework is similar in spirit to the tracking-learning-detection (TLD) system for frame-based cameras \cite{Kalal2012}, nonetheless significantly different in the methods suited to event-based vision. We term our event-based object tracking framework as {\em e-TLD}. This is one of the first works to introduce a general purpose method to track object data from event cameras, which can be efficiently implemented in software at least, in contrast to the ever-growing neural network paradigms that potentially require hours of re-training for online learning. The {\em e-TLD} online learning process is the incremental SVM update stage that is well-documented to be a very efficient process \cite{laskov2006incremental}. Apart from the online learning ability of {\em e-TLD}, the core training process is the codebook learning step that requires under a minute for $500$ms worth of data on a standard PC using efficient sampling strategies \cite{nowak2006sampling}. This also requires significantly lower resources in contrast to Siamese deep neural network object tracking paradigms \cite{DBLP:Siammask} that require ASIC implementations for real-time inference on each video frame. In summary, the objective is to develop a real-time long-term tracking system to achieve: 

\begin{enumerate}
	\item Continuous, long-term robust tracking under background change, illumination change and scale change.
   \item Re-capturing the target after temporarily occluded by other objects or when it re-appears after exiting.
\end{enumerate}


\begin{figure}[!t]
\centering
\subfloat[]{\includegraphics[width=1.8cm]{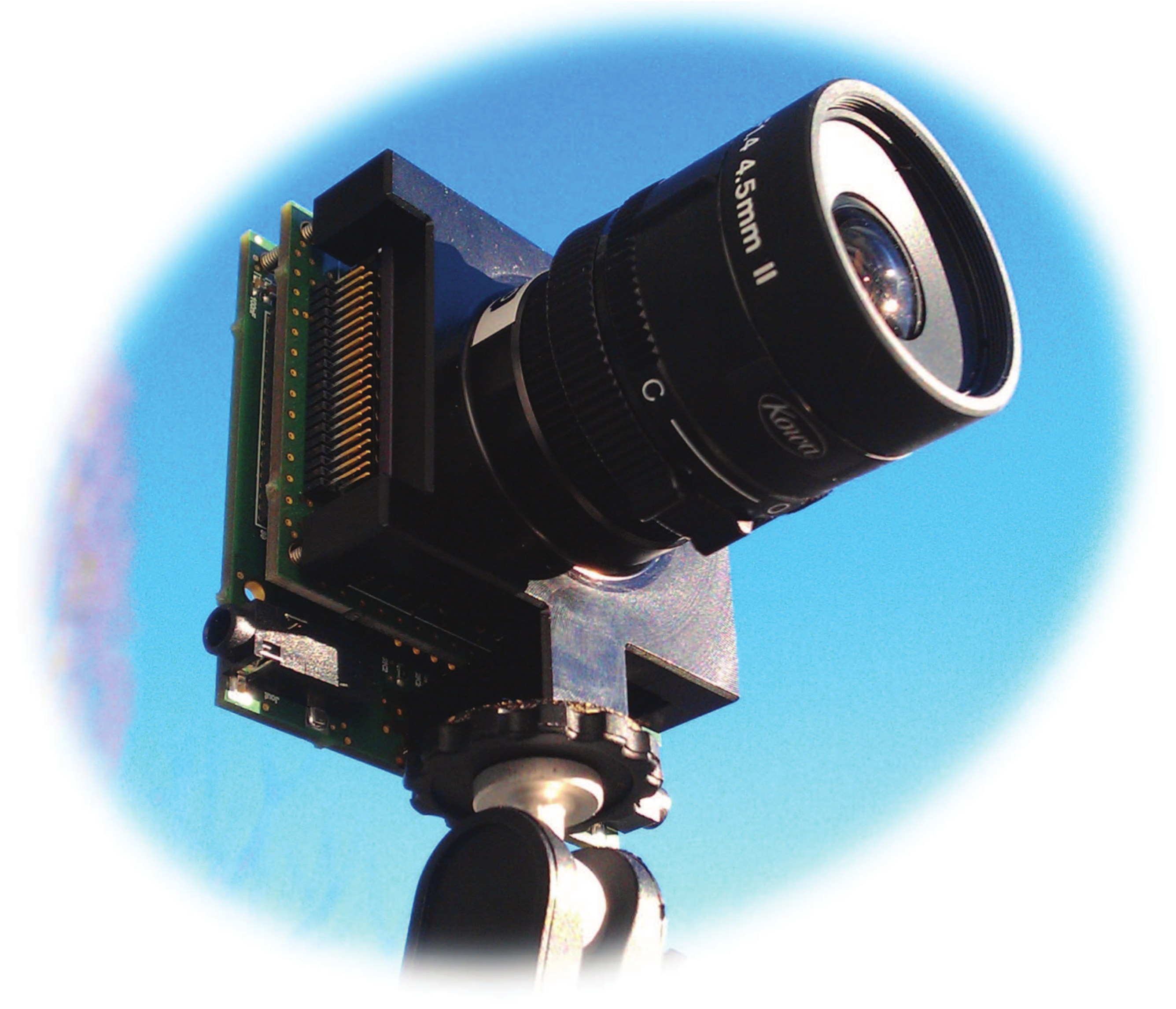}%
\label{fig1_first_case}}
\hfil
\subfloat[]{\includegraphics[width=3.8cm]{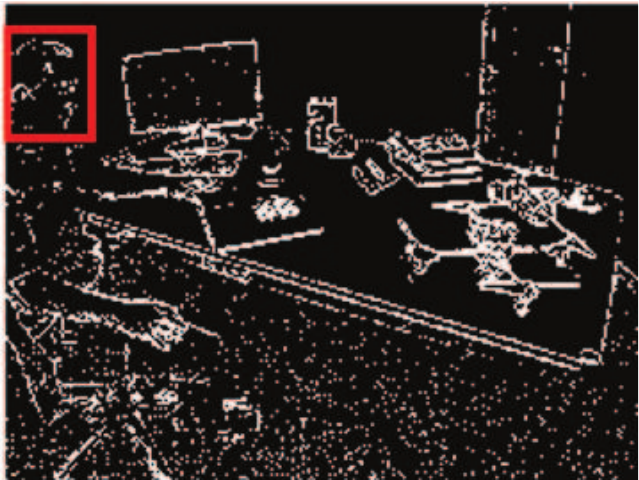}%
\label{fig1_second_case}}
\caption{(a) The DAVIS event camera used in this work (source:inivation.com) (b) The accumulated events shown as a single frame, in which the tracker is initialized with the an arbitrary object position (source: The Event Camera Dataset \cite{Mueggler2017}).}
\label{fig:teaser}
\end{figure}



\section{Event-based Processing}
We use the commercial event camera, the Dynamic and Active-pixel Vision Sensor (DAVIS) \cite{Brandli2014} shown in Fig.~\ref{fig:teaser}(a). It has 240 $\times$ 180 resolution, 130 dB dynamic range and 3 microsecond latency, and communicates with a host computer using USB 2.0. It concurrently outputs a stream of events and frame-based intensity read-outs using the same pixel array. As mentioned earlier in Sec. \ref{sec:intro}, an event consists of a pixel location, a binary polarity value for positive or negative change in log intensity and a timestamp in microseconds. The event camera output can be visualized as shown in Fig.~\ref{fig:teaser}(b) by accumulating events within a short time-period (40ms in this case). In our work, polarity of the events are not considered, so both on-events and off-events are shown in white and the black regions correspond to inactive pixels. Note that only the event data of the DAVIS is used in this work. 
\subsection{Related Work}
\label{subsec:relatedwork}
The recent deep learning revolution in computer vision has also influenced neuromorphic vision with many works primarily in machine learning \cite{Stromatias2017, Lee2016, Neil2016, Ghosh2014, OConnor2013}. Besides learning, simultaneous localization and mapping (SLAM) is a trending robotics application using silicon retinas \cite{Mueggler2014,Rebecq2016,Mueggler2017,Kueng2016,Rebecq2017,Kim2016}. On a larger front, these revolutionary cameras allow new perspectives in reformulating traditional vision problems, such as object detection and tracking, which largely remains an unexplored area of research. 
\par
A few works have used the event camera for object tracking with focus on specific application scenarios. One of the first object tracking applications demonstrated using the commercial dynamic vision sensor (DVS) was to track and control the position of a pencil balanced on a robot arm using a fast event-based Hough transform \cite{Conradt2009}. Other works focused on event-based algorithms for traffic monitoring \cite{Hinz2017, Litzenberger2006} from a static sensor point-of-view, and consequently, tracking can be treated without background modeling as only dynamic objects are picked out by the static event camera. Similarly, the robot goalie application \cite{Delbruck2013} also takes advantage of the stationary DVS camera for tracking multiple balls. Recently, \cite{tracker5_afshar2019eventbased} proposed an event-based algorithm that can perform tasks such as detection and tracking designed specifically for space situational awareness applications. 
\par
A handful of works have attempted to tackle tracking of objects from a moving event camera. Using the DAVIS, \cite{Liu2016, tracking4_CNNcorrelation, tracker2_offon2020object} use a convolutional neural network (CNN) to detect likely target locations for tracking from a moving platform. However, a hybrid approach with frames and events naturally loses the advantages of a low-latency, purely event-driven approach, although can provide energy savings in hardware implementations \cite{tracking1_singla2020hynna}. On the other hand, \cite{tracking3_movingobjIROS2018} uses a parametric model to motion-compensate for the camera, without explicit feature tracking or optical flow computation, and subsequently moving objects that do no confirm to the model are detected in an iterative fashion. Nonetheless, data association and re-detection for long-term tracking remain missing components in these approaches. 
\par
Compared to above works, general purpose object tracking works \cite{Valeiras2015, Lagorce2015} using event-based approaches have been proposed to track incoming blobs of events based on local shape properties. Although these methods are capable of adapting its shape and position to the distribution of incoming events, they carry motion assumptions such as a bivariate Gaussian distribution. Thus, the algorithm parameters were defined experimentally according to the target to track, as acknowledged in \cite{Valeiras2015}. Moreover, the previous systems are not suited to track a set of patterns/object as a whole. Finally, \cite{Valeiras2015, Lagorce2015} are not suited for long-term tracking, because there is no detector to re-initialize the tracker after a failed track. 
\par
In contrast to the above works, the event-based long-term object tracking and detection framework proposed in \cite{Ramesh2019} has the following limitations. As acknowledged in \cite{Ramesh2019}, our prior method works only when there is a clean background surrounding the target object. Secondly, the training phase of the detector \cite{Ramesh2019} is not data-driven (less reliable in practice) and uses computationally expensive image processing approaches for locating the most probable object candidate. Thus, our previous work is suitable only for simple shapes, as shown in the accompanying video results. 
\par
In this paper, we propose a general purpose discriminative tracking system using a \emph{local sliding window} approach, whose parameters are intuitive and can be easily generalized for a wide variety of objects having different shapes and sizes in cluttered settings, as shown in Fig.~\ref{fig:teaser}. The classifier used is a support vector machine (SVM) with an additive $\chi^2$ kernel. For efficient implementation, finite dimensional linear approximations of the kernel are used, as introduced in \cite{Vedaldi2010}. Such maps are efficient linear representations of popular ones, such as the intersection, \(\chi^2\), and Jensen-Shannon kernels. Moreover, with a computationally easier online update, SVM is preferred over other classifiers and deep learning approaches. Lastly, we propose a \emph{data-driven} approach for training the object detector and a \emph{global sliding window} method for locating the object, which allows detecting even small objects, like the drinking cup near the monitor in Fig.~\ref{fig:teaser}(b).

\subsection{Contribution}
\label{subsec:contrib}
This paper is an extended version of the preliminary work  \cite{Ramesh2018}. Novel contributions over \cite{Ramesh2018} include quantitative analysis on the full-length recordings of the event camera dataset \cite{Mueggler2017}, tested for the first time using event-based sensors to the best of our knowledge (Sec.~\ref{subsec:results}), and robustness analysis using hand-held experiments (Sec.~\ref{subsec:realtime}) with critical insights into the system performance for various hyper-parameters (Sec.~\ref{sec:discuss}). We also release full-length annotations for the dynamically captured data, i.e., moving objects captured with a moving camera setting. Additionally, this work includes a comprehensive comparison to existing state-of-the-art event-based tracking method e-LOT \cite{Ramesh2019} (Sec.~\ref{sec:elotcompare}) and further provides new implementation details in Sec.~\ref{sec:methods}, including a free-running mode implementation capable of a detection output at any point in time, as opposed to periodically operating on a set of events \cite{Ramesh2019, Ramesh2018}. Finally, we have tightly integrated the tracker and detector with parameter sharing and in the process also obtain better performance compared to \cite{Ramesh2018}. 

\section{Methodology}
\label{sec:methods}
\begin{figure}%
\centering
\includegraphics[width=3in]{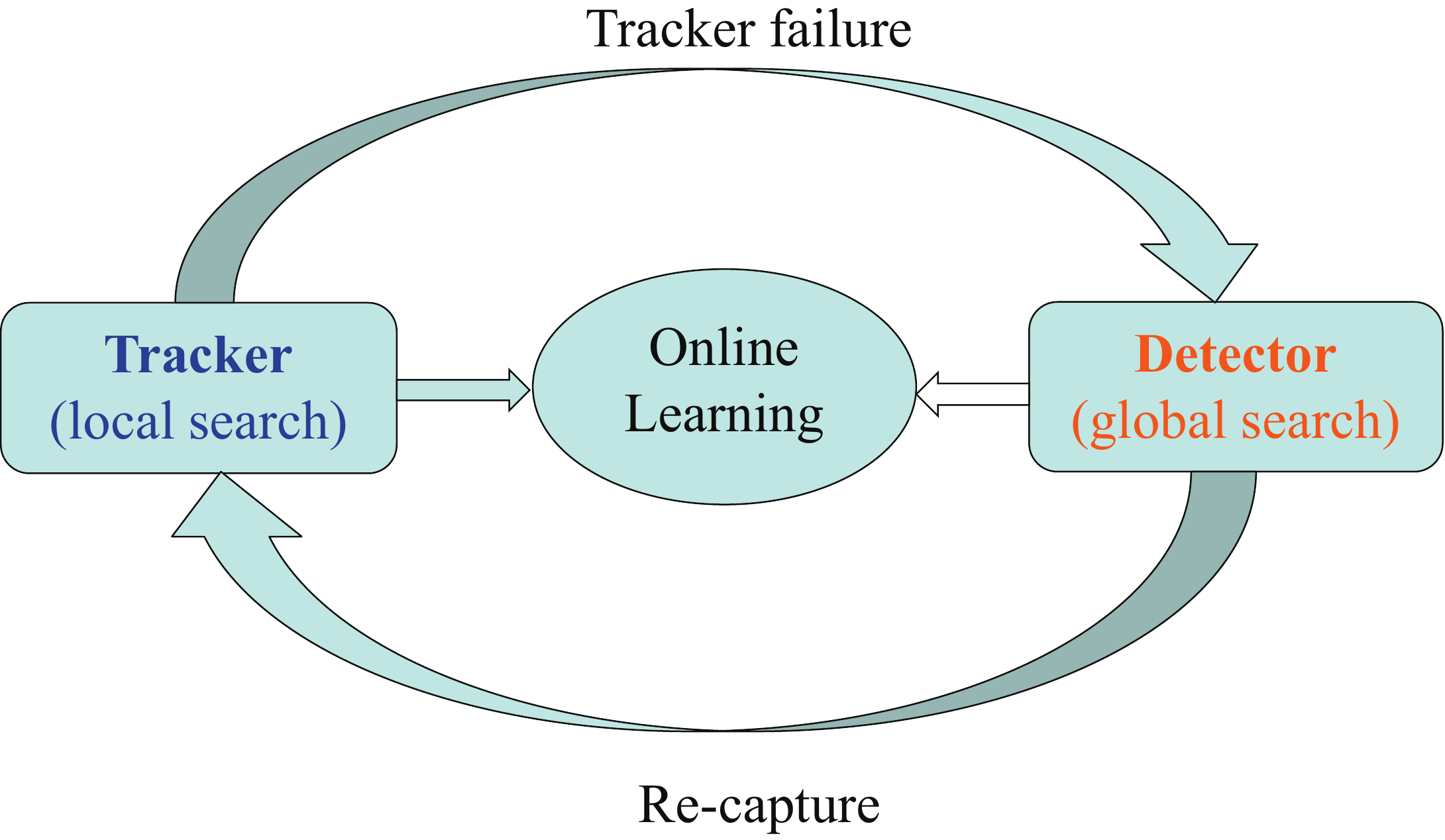}%
\caption{Tracker and detector flow of the proposed {\em e-TLD} framework.}%
\label{fig:etld}%
\end{figure}
The proposed {\em e-TLD} framework integrates a tracker and detector, as shown in Fig.~\ref{fig:etld}. The event-based object tracker (Sec.~\ref{sec:tracker}) is a local search that requires initialization and outputs smooth trajectories. However, it cannot recover from failure on its own. The event-based object detector (Sec.~\ref{sec:detector}), on the other hand, is a global search that does not assume anything about the previous position of the object, and is relatively slower compared to the tracker. However, we can achieve real-time processing by activating the detector only when tracker failure happens. 
\par
During the tracking process, online learning is needed to account for the changes in object appearance. In particular, the binary classifier used by the tracker is updated when the region-of-interest (ROI) is classified as the object. Updating the tracker mitigates the drifting issue, but only done when the tracking confidence is higher than a percentage of the mean tracking score. If tracking failure happens, a higher confidence value is needed to re-activate the tracker. In other words, the target will be re-tracked only when it can pass both the detector and a more ``strict'' tracker. The following subsections describe the {\em e-TLD} framework to jointly track and detect the object. 
\subsection{Event-based object tracker}
\label{sec:tracker}
\begin{figure}[t]%
\centering
\includegraphics[width=2.5in]{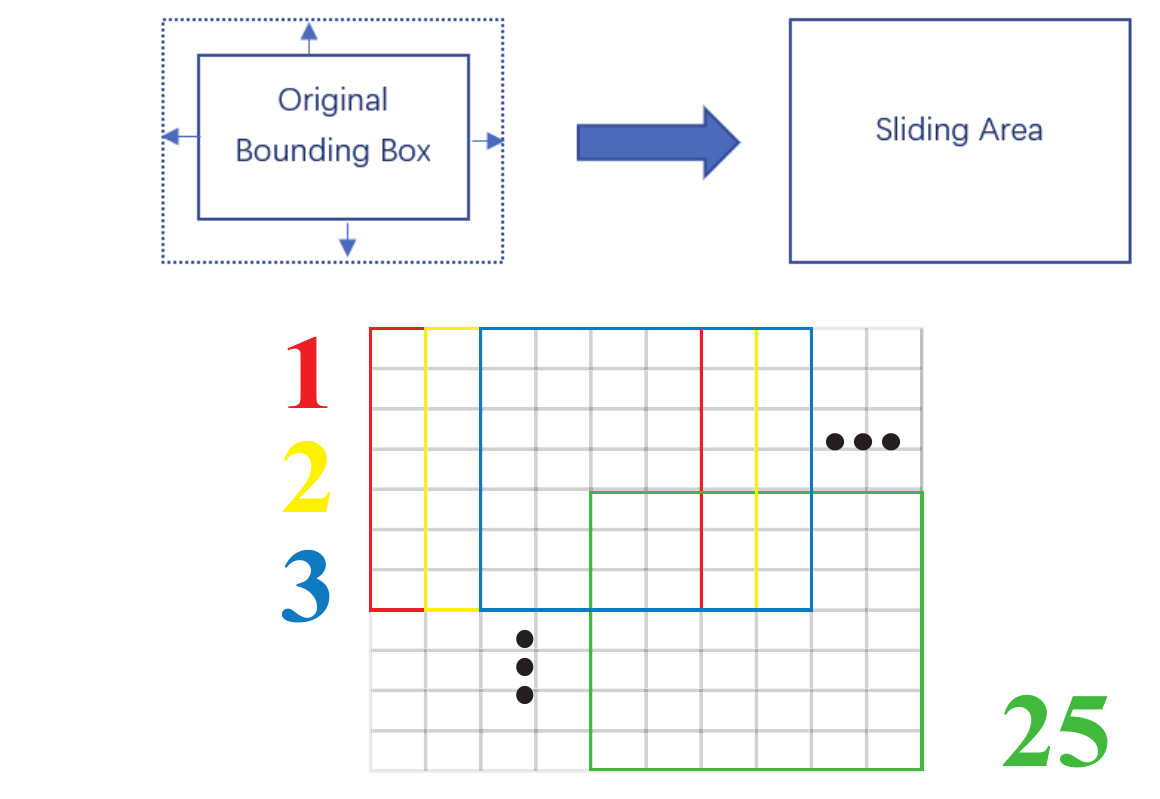}%
\caption{Local sliding window for object tracking using event cameras. A small padding ensures the sliding area contains the object in the next instance of classification. As shown in the example above, a padding of two pixels in $x$ and $y$ directions creates 25 candidate windows (best viewed on monitor).}%
\label{fig:locslide}%
\end{figure}
Each time a ROI is classified as object, a small padding ensures the search area contains the object at the next instance of classification, as shown in Fig.~\ref{fig:locslide}. The position of the object is then updated with the candidate ROI with the highest classification score. This process is extremely simple, but works extremely well in challenging cluttered conditions due to the high-temporal resolution of the event camera. Note that we set the classification period in terms of the number of events received within the ROI, instead of explicitly choosing a time-period. In particular, this number is chosen as a small fraction of the bounding box size and thus allows a dynamic classification rate for different object shapes and sizes. 
\par
We employ the feature descriptor proposed in \cite{Ramesh2017a}, and thus, each event is encoded as a local descriptor. The notation ${\bf{e}}_i = (x_i,y_i,t_i,p_i,{\bf{x}}_i^T)^T$ denotes an event with pixel location $x_i$ and $y_i$, timestamp $t_i$, polarity $p_i$ and the feature vector ${\bf{x}}_i$. 
\par
We denote by $N$ the number of candidate windows, and by $X^j = ({\bf{x}}_1, {\bf{x}}_2, \cdots, {\bf{x}}_{n_i})$ the collection of event descriptors contained within a candidate window $W^j$ where ${\bf{x}}_l \in \mathbb{R}^d$, $l=1,2,\cdots, n_i$ is a descriptor in feature space $S$.  
\par
Inspired by the bag-of-words (BOW) model in computer vision \cite{Csurka04visualcategorization}, each feature vector ${\bf{x}}_l$ is quantized into one of $K$ different visual words that are obtained from the training phase. The mapping to a visual word $v_k \in S$ is achieved using a quantization function $f_k({\bf{x}})~:~ S \mapsto \{0,1\}$. Each quantization function $f_k({\bf{x}})$ is essentially computing the distance of the feature vector to $v_k$ and allowing the assignment if it is minimal. 
\begin{equation}
f_k({\bf{x}}) = f({\bf{x}}; v_k) = I(||{\bf{x}} - v_k|| = \rho)
\label{eq:vectquant}
\end{equation}
where indicator function $I(z)$ outputs 1 when $z$ is true or 0 otherwise; $\rho$ is the Euclidean distance, ${\arg\min_{k}} ||{\bf{x}} - {v_k}||$ . Given K visual words, or K quantization functions $\{f_k({\bf{x}})\}_{k=1}^K$, a codeword representation is computed as,
\begin{equation}
h_j^k = \frac{1}{{{n_i}}}\sum\limits_{l = 1}^{{n_i}} {{f_k}({\bf{x}}_l)} 
\label{eq:bow}
\end{equation}
The tracker representation for $W^j$ is expressed by the vector,
\begin{equation}
 {\bf{h}}_j=(h_j^1,h_j^2,\cdots,h_j^K)
\label{eq:BOW}
\end{equation}
Each incoming event in a candidate window $W^j$ is then used to update the tracker representation ${\bf{h}}_j \in \mathbb{R}^K$. The scalar-valued discriminant function $D({\bf{h}}_j)$ indicates the presence (class $\omega _1 \Rightarrow{+1}$) or absence of the object (class $\omega _2\Rightarrow{-1}$) dynamically, 
\begin{equation}
D({\bf{h}}_j) \gtrless 0 \Rightarrow {\bf{h}}_j 
\label{eq:svmfunc}
\end{equation}
where ${h}_j \in \{\omega_1, \omega_2\}$.
During the training phase, where the user specifies a tight ROI in space-time that contains the object, all the events including ones outside the ROI are used to obtain the parameters of $D$, which is the problem of constructing a classifier for two classes -- object vs background.

\subsubsection{Training phase}
\label{sec:trainclass}
When the user specifies the spatio-temporal position of the object, the first step is create the visual words $\{v_k\}_{k=1}^K \in S$, which are the cluster centers generated using k-means clustering of the event descriptors inside and outside the ROI. In other words, the codebook $C = [v_1, v_2, \cdots, v_K] \in \mathbb{R}^{d \times K} $ is an unsupervised learning step. Then the events within the ROI, represented by the set of descriptors $X^{{\omega _1}} = ({\bf{x}}_1, {\bf{x}}_2, \cdots, {\bf{x}}_{C_1})$ can be used to generate a tracker representation ${\bf{h}}_{\omega _1}$, given by eq. \eqref{eq:BOW}. Similarly, the events outside the ROI $X^{{\omega _2}} = ({\bf{x}}_1, {\bf{x}}_2, \cdots, {\bf{x}}_{C_2})$ can be used for obtaining ${\bf{h}}_{\omega _2}$. However, training a classifier with just one sample from each class (${\bf{h}}_{\omega _1}$ and ${\bf{h}}_{\omega _2}$) is pointless. 
\par
To solve the low sample problem, statistical bootstrapping \cite{Mooney1993} can be used to generate new subsets of descriptors $\{ X^{{\omega _1}}_1, X^{{\omega _1}}_2, \cdots, X^{{\omega _1}}_{n_1}\}$ and $\{ X^{{\omega _2}}_1, X^{{\omega _2}}_2, \cdots, X^{{\omega _2}}_{n_2}\}$. Specifically, bootstraping $X^{{\omega _1}}$ is the process of random sampling of a subset out of the $C_1$ descriptors belonging to the ROI, one at a time such that all descriptors have an equal probability of being selected, i.e., $1/{C_1}$. 
\par
However, storing a set of events or descriptors for bootstrapping ($X^{{\omega _1}}$ and $X^{{\omega _2}}$) is impractical for online learning on an event-by-event basis \cite{Ramesh2019}. Thus, we propose bootstrapping to be interpreted in a Bayesian framework \cite{Rubin1981} that re-weights the histogram representations (${\bf{h}}_{{\omega _1}}$ and ${\bf{h}}_{{\omega _2}}$). Let $P \sim U([0,1])$ be a uniformly distributed random variable. Mathematically,
\begin{equation}
{h^k_{1{\omega _1}}} = \lfloor P \times h_{{\omega _1}}^k \rfloor
\label{eq:bayesbtstrp}
\end{equation}
where the above clipping operator is a floor operation.
Thus, the first bootstrapped histogram representation for the ROI is expressed by the vector,
\begin{equation}
 {\bf{h}}_{1{\omega _1}}=({h^1_{1{\omega _1}}},{h^2_{1{\omega _1}}},\cdots,{h^K_{1{\omega _1}}})
\label{eq:newBOW}
\end{equation}
It is to be noted that eq. \eqref{eq:newBOW} is not a true bootstrap procedure since the maximum values of ${h^k_{1{\omega _1}}}$ need not be clipped to the corresponding maxmimum values of $h_{{\omega _1}}^k$, as seen in eq. \eqref{eq:bayesbtstrp}. However, the Bayesian bootstrap is operationally and inferentially similar to the true boostrap \cite{Rubin1981}. Let $N_1$ and $N_2$ denote the number of samples after bootstrapping belonging to  class $\omega _1$ and $\omega _2$ respectively. Then, the collection of the bootstrapped representations $\{ {\bf{h}}_{1{\omega _1}},{\bf{h}}_{2{\omega _1}}, \cdots , {\bf{h}}_{{N_1}{\omega _1}} \}$ and $\{ {\bf{h}}_{1{\omega _2}},{\bf{h}}_{2{\omega _2}}, \cdots , {\bf{h}}_{{N_2}{\omega _2}} \}$ is used to train the SVM classifier $D(\cdot)$ with a $\chi^2$ kernel \cite{Vedaldi2010}. 
\subsubsection{Tracking Phase}
\label{subsubsec:traintracker}
The candidate windows $\{W^j\}_{j=1}^N$ each output a tracker representation ${\bf{h}}_j$. The best candidate window is chosen to be the tracker state $B_t$ when $D({\bf{h}}_j)$ is maximized. 
\begin{equation}
\underset{X^j \in S}{\operatorname{arg\,max}}\, D({\bf{h}}_j(X^j)) := \{X^j, j = 1, 2, \cdots, N \}, 
\label{eq:svmmax}
\end{equation}
given for all $Y^j$, the discriminant function satisfies $D({\bf{h}}_j(Y^j)) \le D({\bf{h}}_j(X^j))$. The number of events for the ROI update (``waiting time'' between two instances of classification) is set as $\tau \times height \times width$ of the ROI $X^j$, where $\tau \in [0,1]$ is set to $0.05$ in our experiments. Thus, when the sliding area contains 5\% of the events relative to the number of pixels contained within the ROI, the next instance of classification is triggered (see Fig. \ref{fig:locslide}). The average SVM score after several instances is used to determine whether the next instance of tracking is successful. In case, the SVM score falls below a fraction of the average score, $\tau_t$, then the object detector is instantiated to globally search for the object. 
\begin{figure}%
\centering
\includegraphics[width=3.3in]{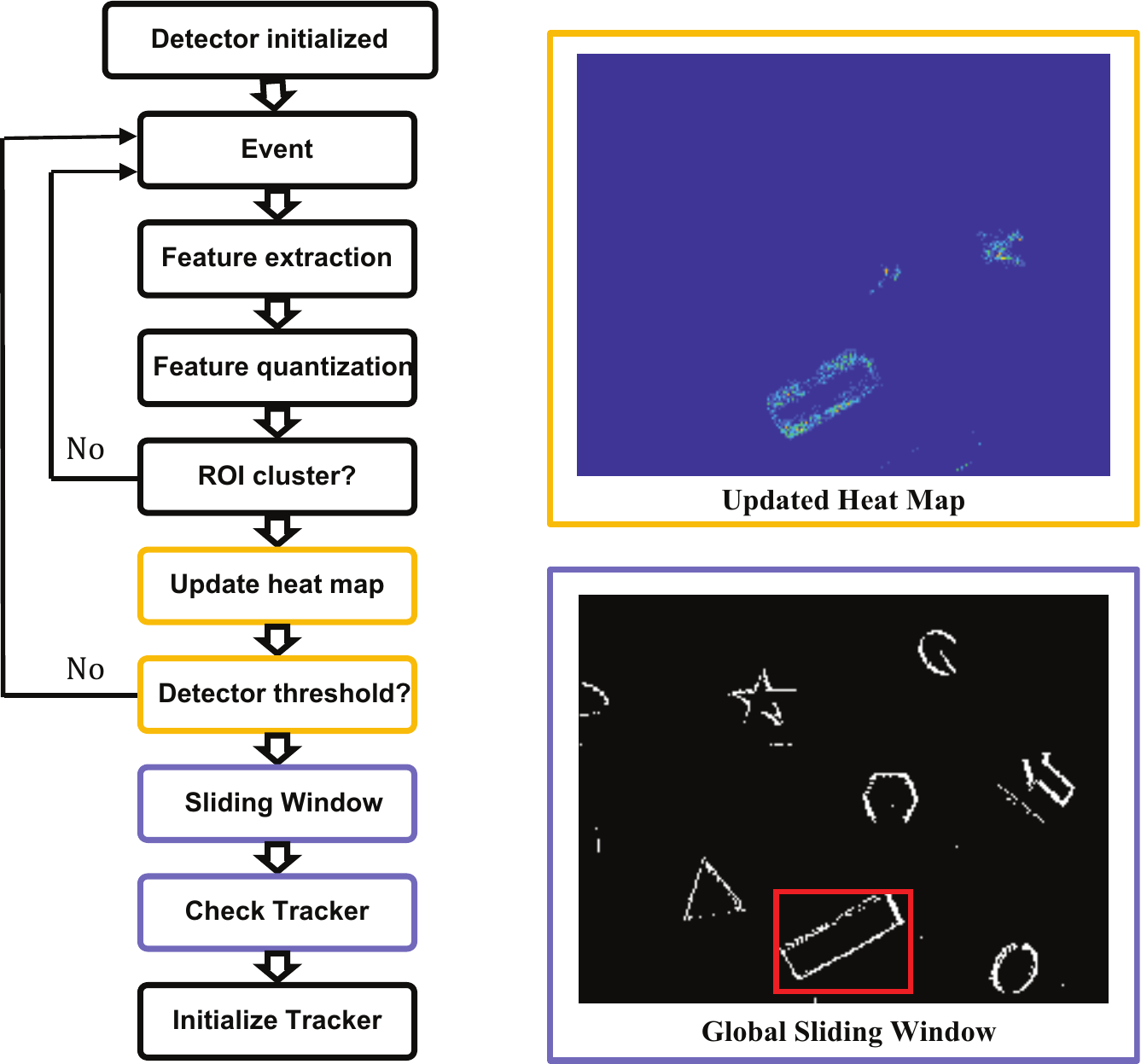}%
\caption{Detector flowchart of the proposed {\em e-TLD} framework.}%
\label{fig:etld_det}%
\end{figure}

\subsection{Event-based object detector}
\label{sec:detector}
Once the tracker has lost the object, detecting the object is the problem of obtaining a candidate ROI and continuing the tracking process. Therefore, detection is a global sliding window search compared to the local sliding window search of the tracker. Fig.~\ref{fig:etld_det} illustrates the detection process that is described in detail below. 

\subsubsection{Training phase}
Similar to the training phase of the tracker (Sec. \ref{subsubsec:traintracker}), the object detector uses the ROI initialization by the user. Let $o$ denote the number of quantized clusters to which the object samples $X^{{\omega _1}}$ are frequently mapped, and the corresponding cluster indices be $O_{{\omega _1}} = \{k_1,k_2, \cdots, k_o\} ~\text{where}~ o \ll K$. The objective of the proposed detector training phase is to obtain $O_{{\omega _1}}$ in a data-driven fashion without relying on ad-hoc threshold parameters. 
\par
The main idea is to deduce clusters that are important to $X^{{\omega _1}}$ while rejecting quantization results that are common to $X^{{\omega _1}}$ and $X^{{\omega _2}}$. By making use of the Bayesian bootsrapped representations, a new vector ${\bf{h}}^{\text{diff}}_{{\omega _1}{\omega _2}} \in \mathbb{R}^{K}$ is used to obtain object clusters for the detection process, 
\begin{equation}
{\bf{h}}^{\text{diff}}_{{\omega _1}{\omega _2}} = \sum\limits_{l = 1}^{{N_1}} {\bf{h}}_{l{\omega_1}}  - \sum\limits_{m = 1}^{{N_2}} {\bf{h}}_{m{\omega_2}}
\label{eq:objdettrain}
\end{equation}
The positive values in $ {\bf{h}}^{\text{diff}}_{{\omega _1}{\omega _2}}$ represent codewords that have been assigned to the object more times than it has been assigned to background. Therefore, these codewords are simply chosen to be $O_{{\omega _1}}$. This data-driven approach of training the detector ensures that the ROI events have the highest probability of detection compared to choosing cluster centers that have a high percentage of ROI events, as was done in \cite{Ramesh2019}. In other words, cluster centers that are selected as detection landmarks do not ensure ROI events belong to the codewords.
\subsubsection{Detection phase}
\label{subsec:detectionphase}
Algorithm~\ref{ag1} outlines the proposed event-based object detection approach. If the event camera output contains $h$ rows and $w$ columns, a detection matrix $M \in \mathbb{R}_+^{h \times w}$ is used to keep track of events that may belong to the object. For every incoming event, the quantization function, defined in eq. \eqref{eq:vectquant}, determines whether it belongs to the detector clusters $\{k_1,k_2, \cdots, k_o\}$ and the corresponding location of the event is used to increment $M$. The detector threshold $\tau \times h \times w$, determines if enough events have been accumulated within the detection matrix $M$, and represent a percentage of the pixels from the ROI. The parameter $\tau$ is the same as the one for the local search tracker, set to 0.05, meaning at least 5\% of the events have occurred globally for the detection process. 
\par
A global sliding window process is then performed on $M$ to determine the region with maximal activation due to the presence of the object (if any). If the previous successful object state $B_t$ has m rows and n columns, then the size of the activation map after the global sliding window operation will be $h - m + 1$ rows, and $w - n + 1$ columns.  
\begin{equation}
O(r,s) = \sum\limits_{k = 1}^m {\sum\limits_{l = 1}^n {M(r + k - 1,s - l + 1)} }
\label{eq:convobj}
\end{equation}
where $r$ and $s$ varies from $1,2,\cdots,h - m + 1$ and $1,2,\cdots,w - n + 1$ respectively. The region centered at $max(O(r,s))$ is the new tracker state $B_t$. 

\begin{algorithm}[t]
\caption{Event-based Object Detection}
\label{ag1}
\textbf{Input}: Image size ($h \times w$), detection matrix $M = 0_{h,w}$, threshold $\tau$, $\text{count} =0$, Codebook size $K$\\
\textbf{Output}: Estimated tracker state $B_t$ with ROI width $n$ and height $m$
\begin{algorithmic}[1]
\State{For each incoming event ${\bf{e}}_i =(x_i,y_i,t_i,p_i,{\bf{x}}_i^T)^T$}
\For{$k=1:K$}
\State{Get $f_k({\bf{x}}_i)$ using eq. \eqref{eq:vectquant}}
\If{$f_k({\bf{x}}_i) = 1$ and $k \in O_{\omega_1} = \{k_1,k_2, \cdots, k_o\} $}
\State{$M(y_i, x_i) = M(y_i, x_i) + 1$}
\State{$\text{count} = \text{count} + 1$}
\EndIf
\EndFor
\If{($\text{count} > \tau \times h \times w $)}
\State{Global sliding window eq. \eqref{eq:convobj}:}
\State\hspace{\algorithmicindent}{$O \in \mathbb{R}_+^{{(h - m + 1)} \times {(w-n+1)}}$}
\State{Choose highest activation in $O$ to re-initialize $B_t$}
\EndIf
\end{algorithmic}
\end{algorithm} 
In the case of dynamic objects, the detection matrix $M$ accounts for both the camera motion and object motion. This results in a trail of object events rather than a crisp detection as shown in the heat map of Fig.~\ref{fig:etld_det}. In these cases, our previous system \cite{Ramesh2018} detected very large bounding boxes around the object due to having a different threshold ($\tau_d$ set as 0.25) for the detector that decoupled its behavior with the tracker. This is a seemingly innocuous issue, but one that results in heavy performance loss as shown in the experiments. In this work, the parameter $\tau$ is shared by the tracker and detector, which tightly couples their overall performance. 
\subsection{e-TLD}
\label{sec:eTLD}
As shown in Fig.~\ref{fig:etld}, the event-based tracking-learning-detection ({\em e-TLD}) framework combines the tracker (Sec.~\ref{sec:tracker}) and the detector (Sec.~\ref{sec:detector}) to track a desired object indefinitely. This novel framework integrates tracking and detection together and benefits one from the other to solve the long-term tracking problem. 
\par
There are three advantages for our proposed framework. First, since the global sliding window update of the detector is relatively time-consuming compared to the local sliding window update of the tracker, it is activated only when the local search tracker fails, which reduces the computational complexity significantly. Second, the robustness of the long-term tracker is benefited from treating normal tracking and recovery from detection independently. In particular, when the detector outputs a candidate location, the tracker confidence needs to be above the mean tracking score ($\tau_t =1$) instead of a fraction of it ($\tau_t < 1$). Third, drifting on the tracker is prevented by only updating when the current tracker score is greater than the mean tracking score. 
\par
The main premise in tackling object appearance changes is that the tracker representation (eq. \eqref{eq:BOW}) obtained using the spike context descriptor \cite{Ramesh2017a} is robust to gradual scale and rotation changes. Specifically, the spike context descriptor uses a log-polar grid that tolerates moderate scale and rotation variations. Therefore, as with the case of object tracking scenarios using event cameras with high temporal resolution, the change in appearance from one instance of tracking to another is smooth and thus online learning enables accurate tracking. In the current {\em e-TLD} setup, the detector is not updated on-the-fly, as it requires online dictionary learning, which remains a future direction of research.


\begin{figure*}[!t]
\centering
\subfloat[Tracked object 16.33\% (Total events:15078)]{\includegraphics[width=5cm]{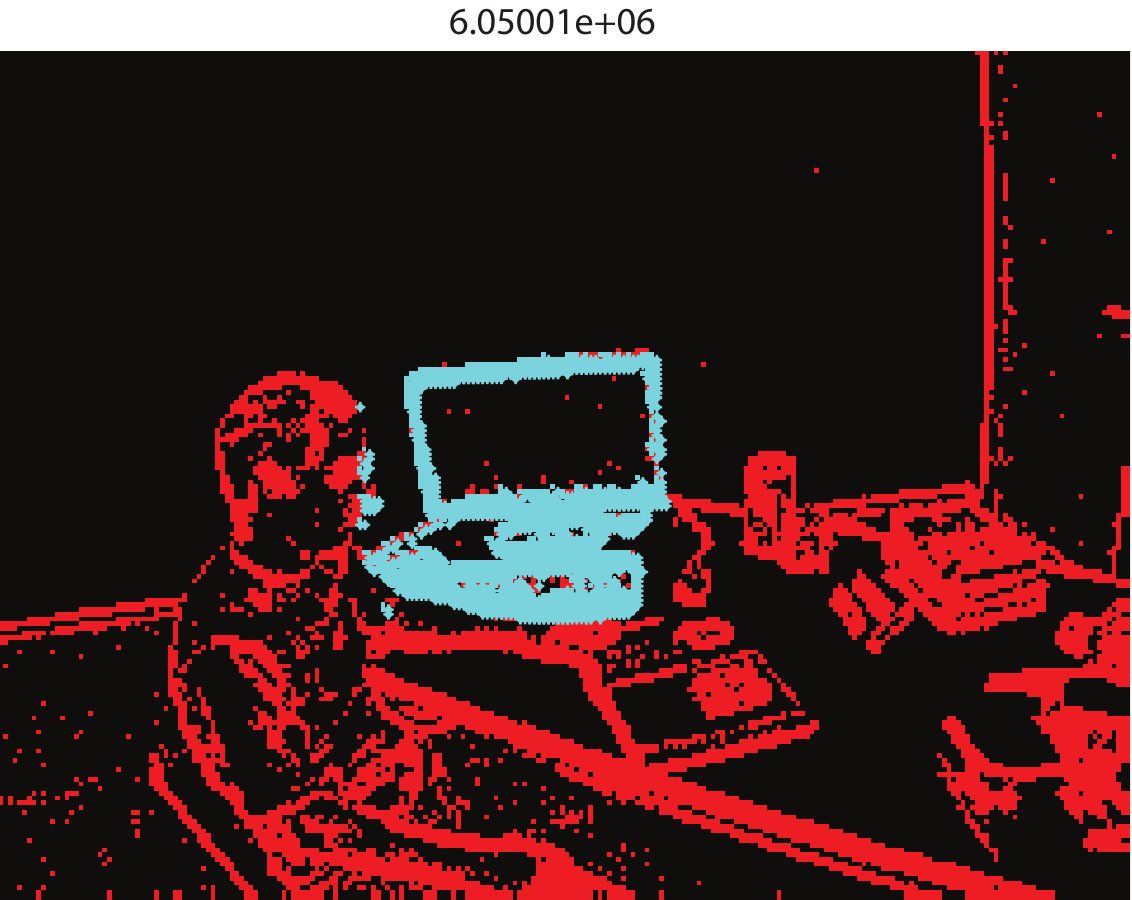}%
\label{fig4_first_case}}
\hfil
\subfloat[Tracked object 10.75\% (Total events:11927)]{\includegraphics[width=5cm]{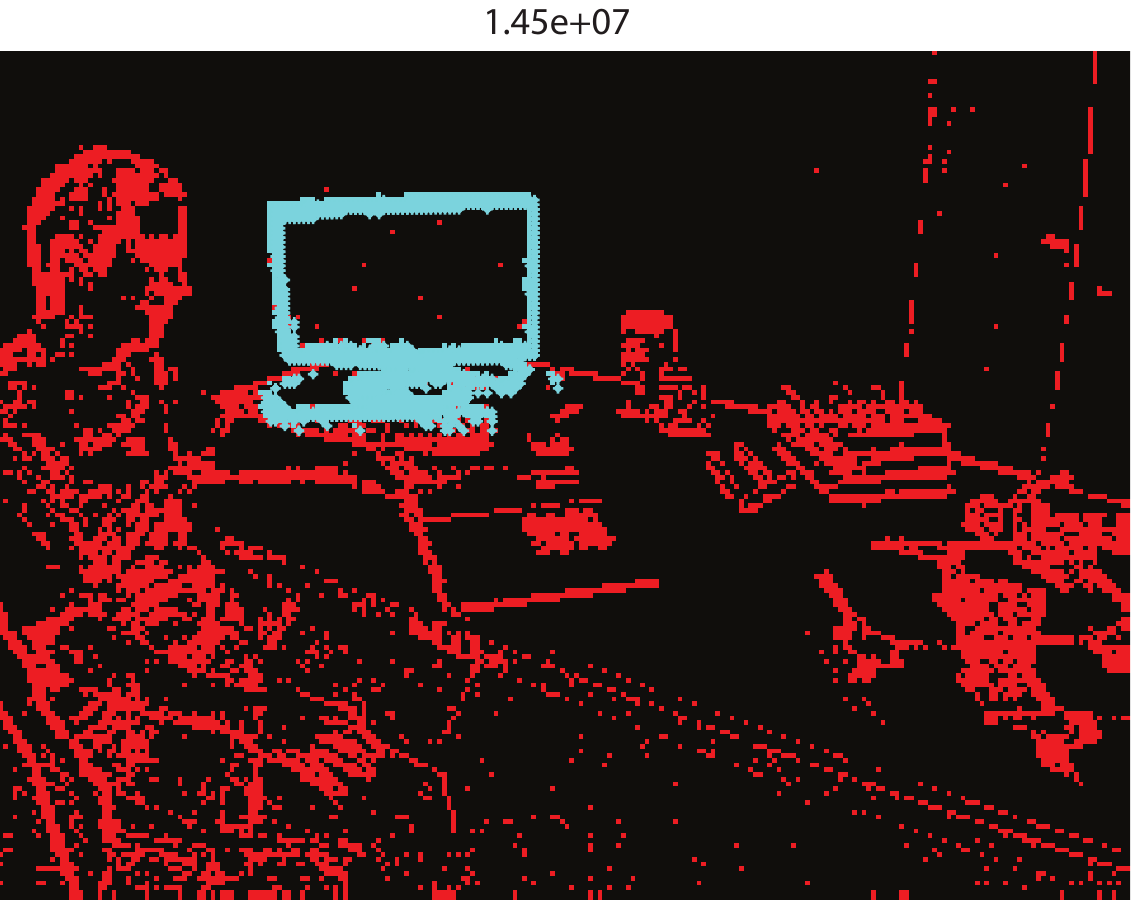}%
\label{fig4_second_case}}
\hfil
\subfloat[Tracked object:8.55\% (Total events:9327)]{\includegraphics[width=5cm]{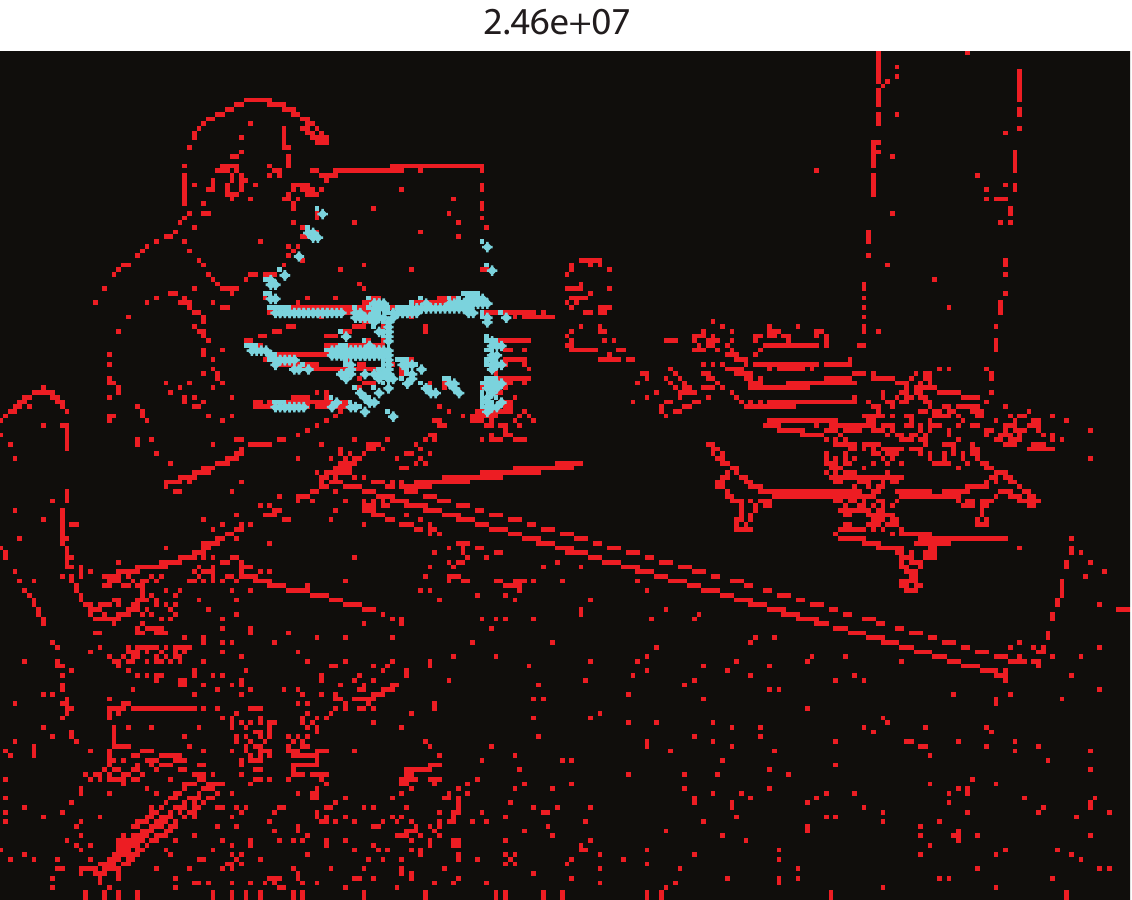}%
\label{fig4_third_case}}
\hfil
\subfloat[Tracked object:3.84\% (Total events:7540)]{\includegraphics[width=5cm]{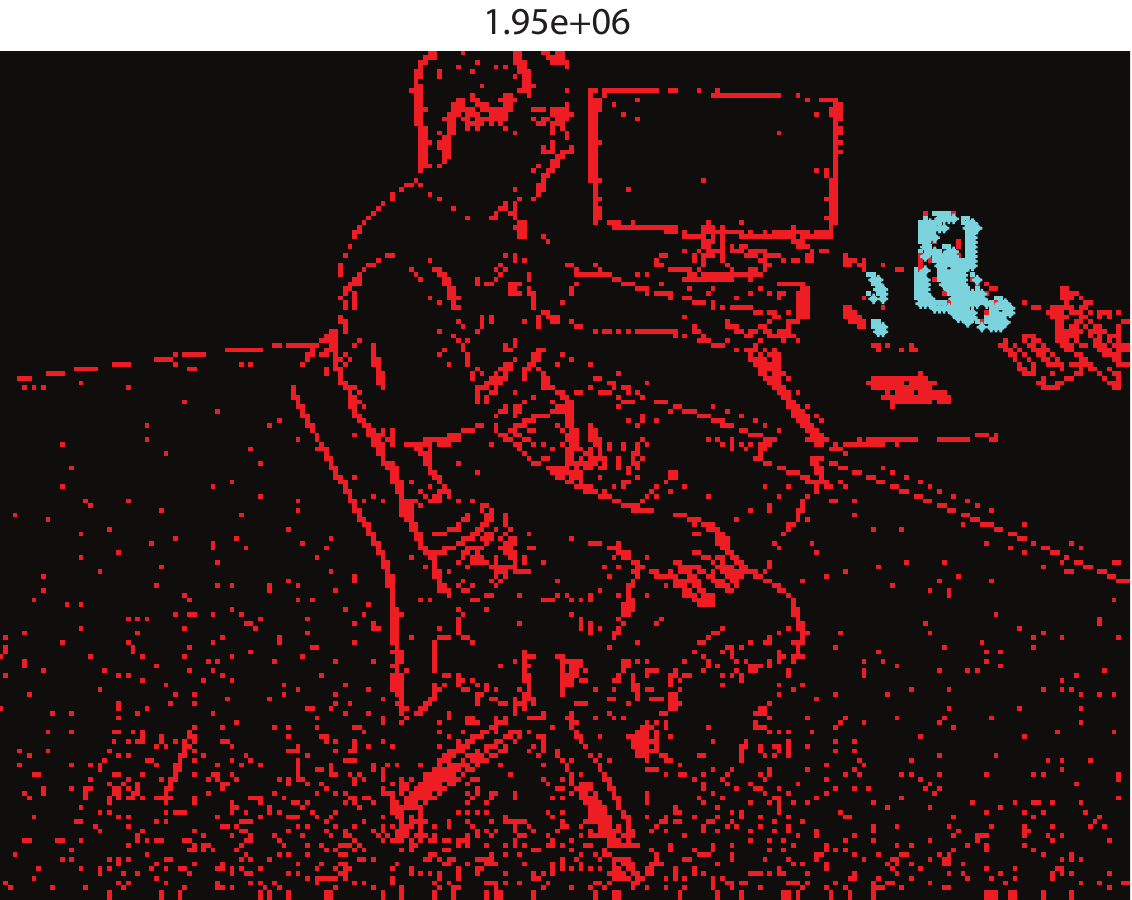}%
\label{fig4_seven_case}}
\hfil
\subfloat[Tracked object:2.59\% (Total events:10782)]{\includegraphics[width=5cm]{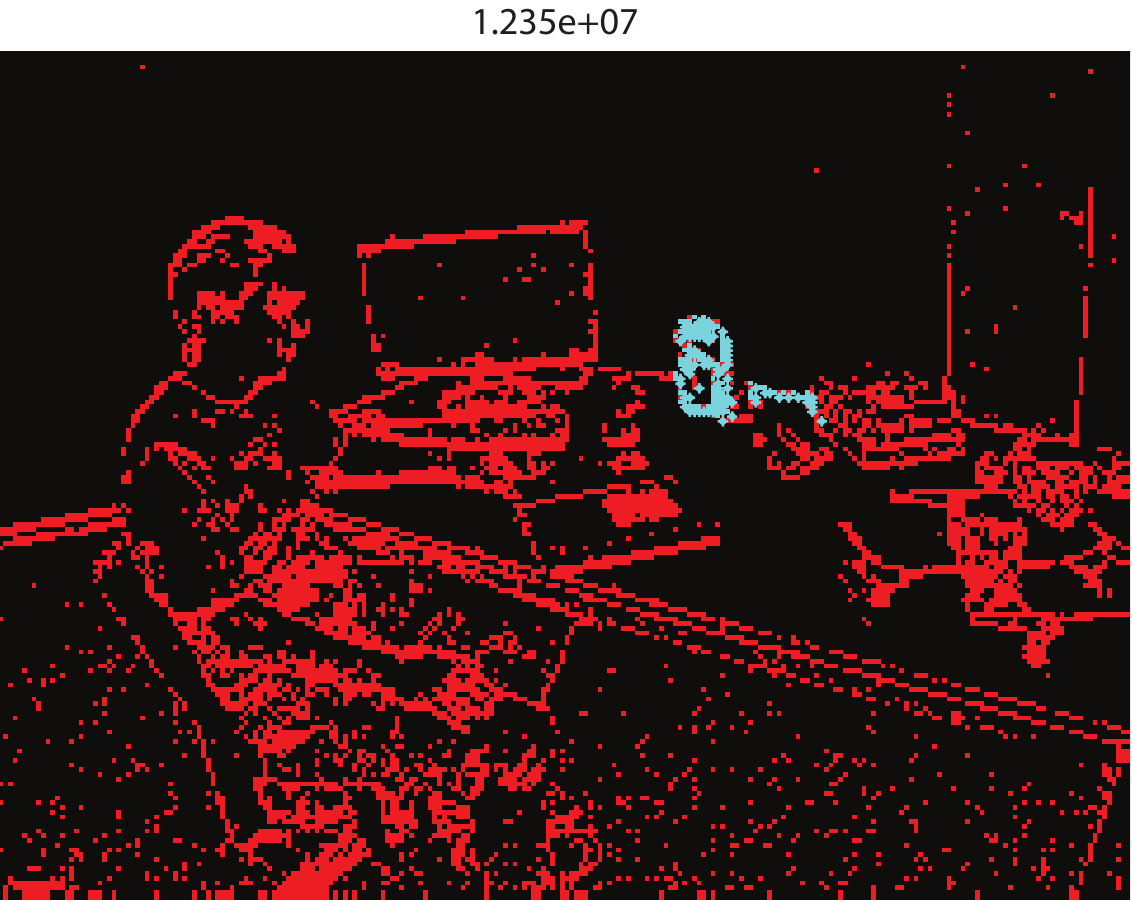}%
\label{fig4_eight_case}}
\hfil
\subfloat[Tracked object:2.15\% (Total events:9317)]{\includegraphics[width=5cm]{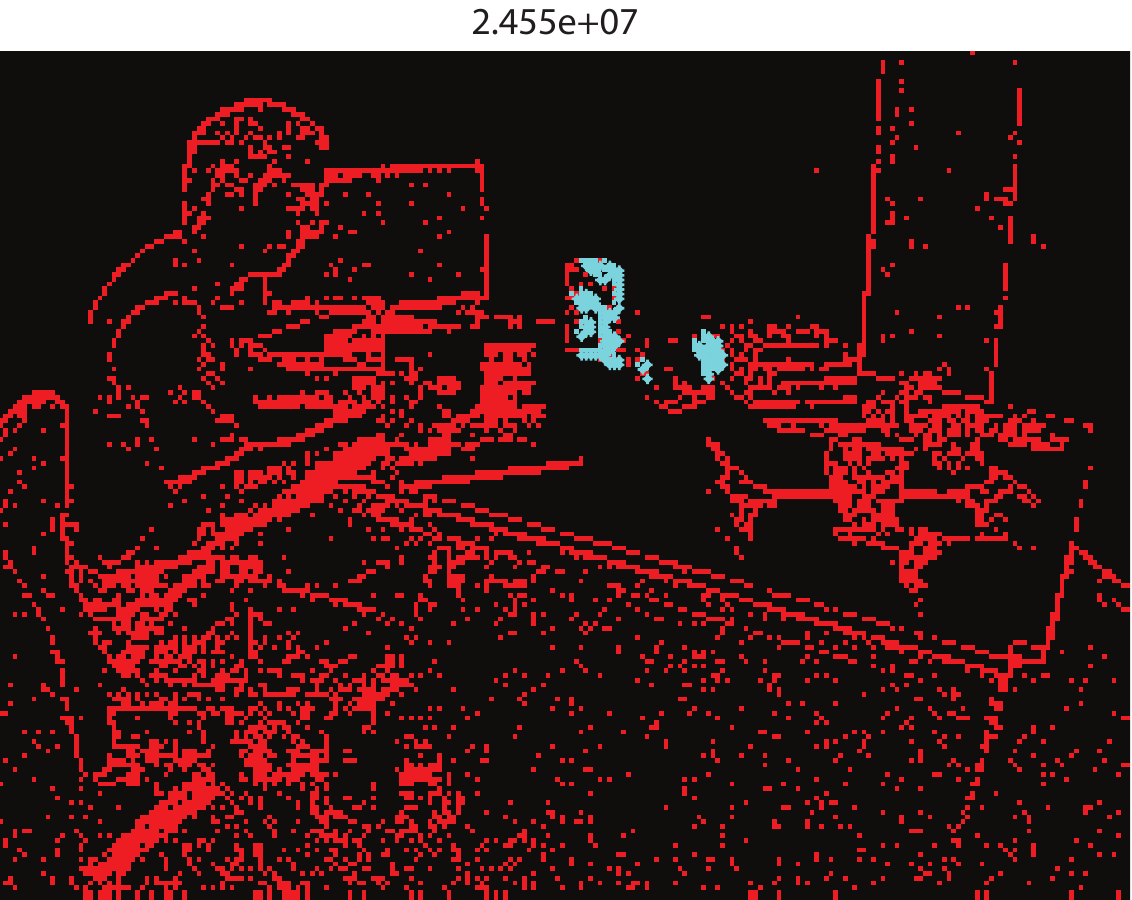}%
\label{fig4_nine_case}}
\caption{Tracking of objects having different shapes and sizes. Each row shows the tracking of a single object. The marked events represent the tracked position in the field-of-view of the event camera and the title of each subfigure displays the time ($\mu$s) instance of the track.}
\label{fig:trackingresults}
\end{figure*}

\begin{figure*}[!t]
\begingroup
\captionsetup[subfigure]{font=scriptsize,labelfont=scriptsize}
\subfloat[Tracked:20.2\% Total:10703]{\includegraphics[width=3.5cm]{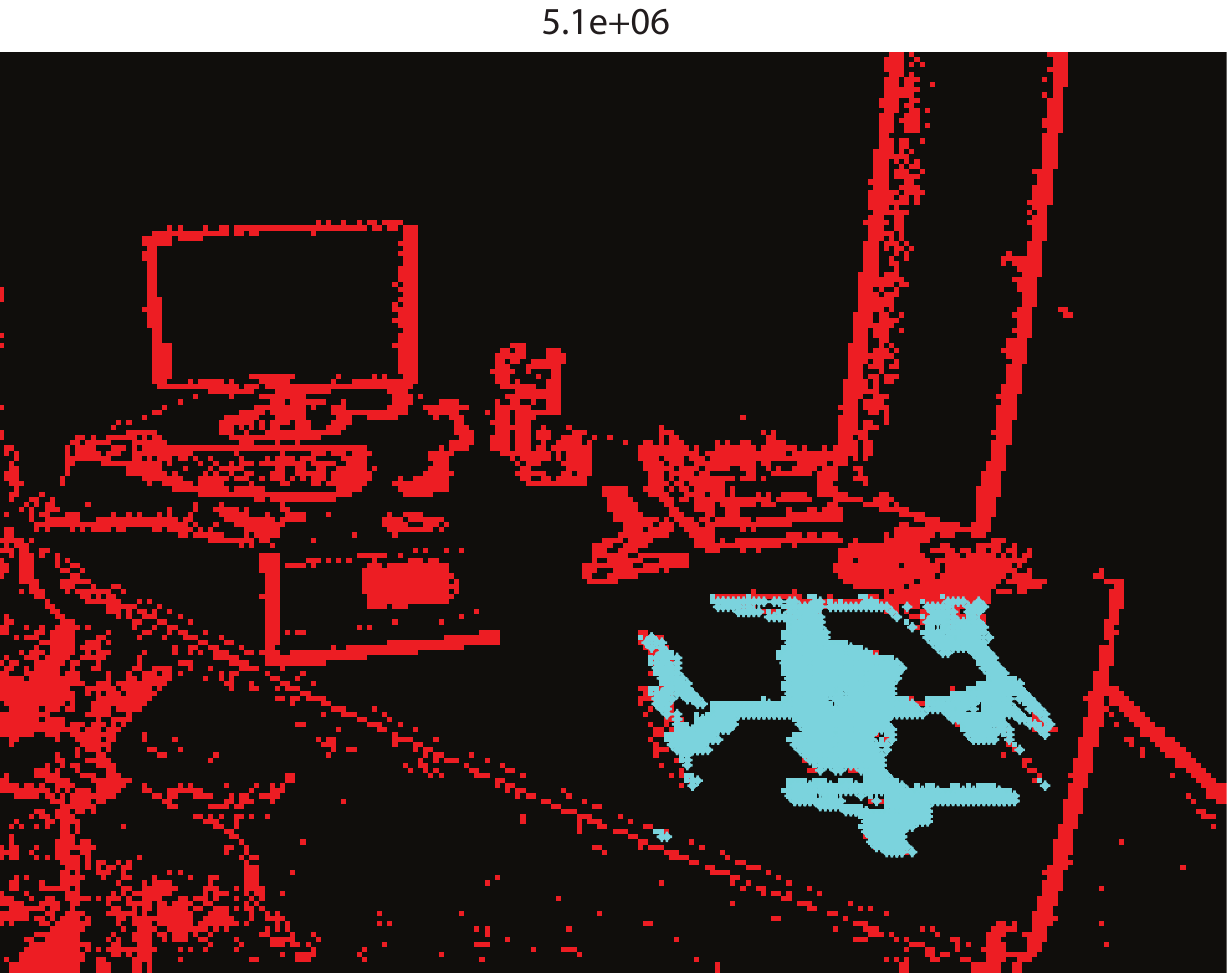}%
\label{fig5_first_case}}
\hfil
\subfloat[Tracked:0\% Total:15624]{\includegraphics[width=3.5cm]{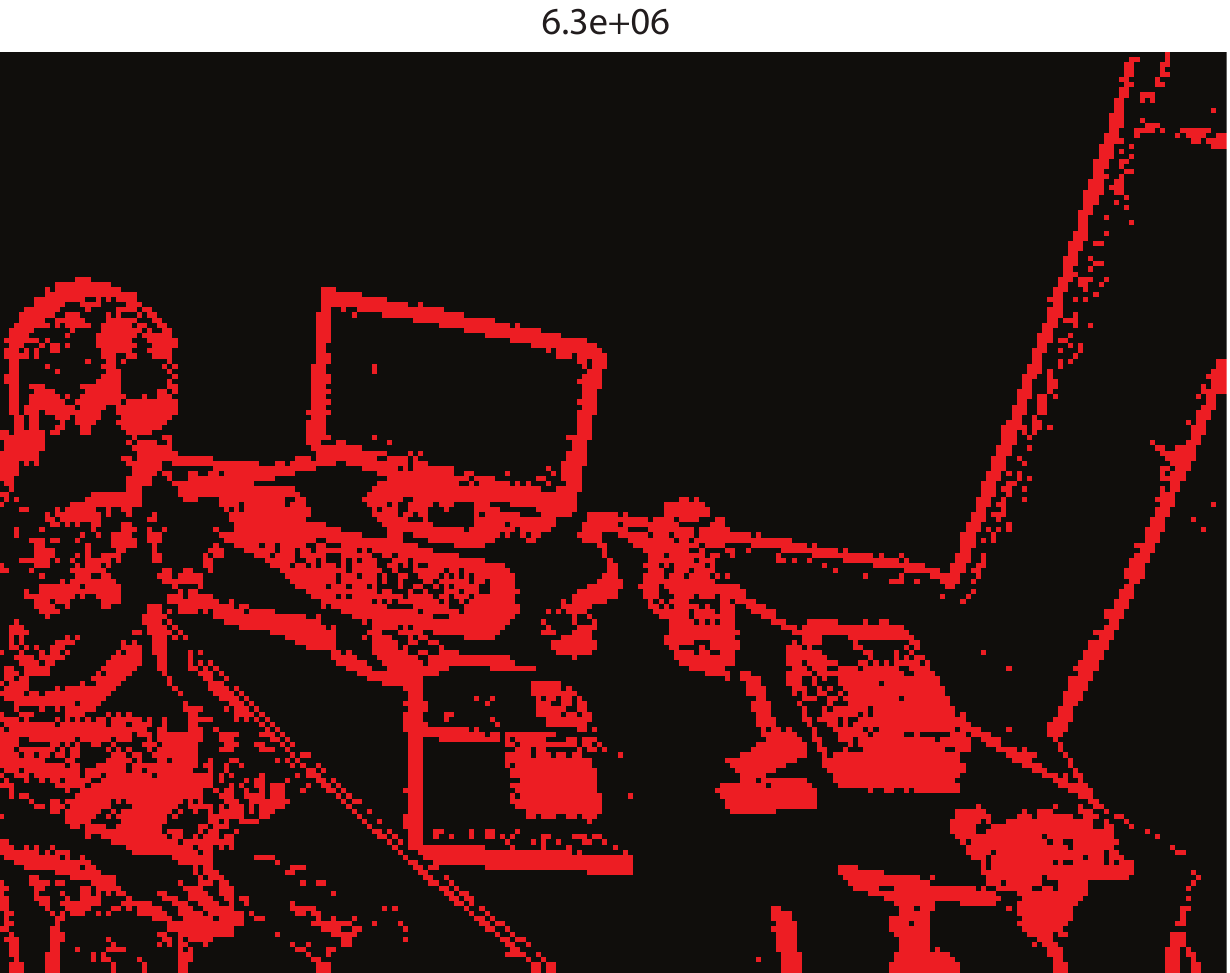}%
\label{fig5_second_case}}
\hfil
\subfloat[Tracked:14.4\% Total:13984]{\includegraphics[width=3.5cm]{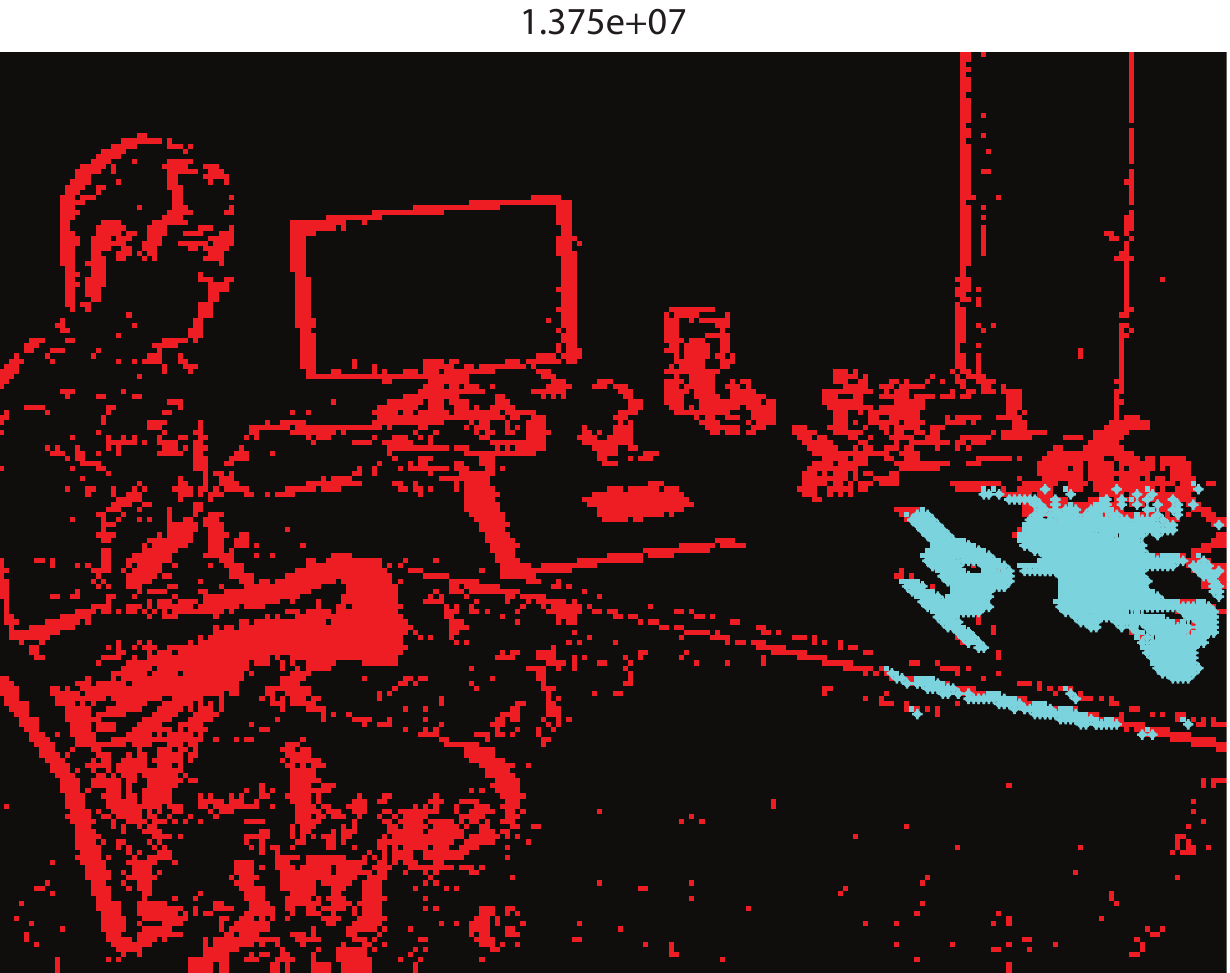}%
\label{fig5_third_case}}
\hfil
\subfloat[Tracked:17.76\% Total:8871]{\includegraphics[width=3.5cm]{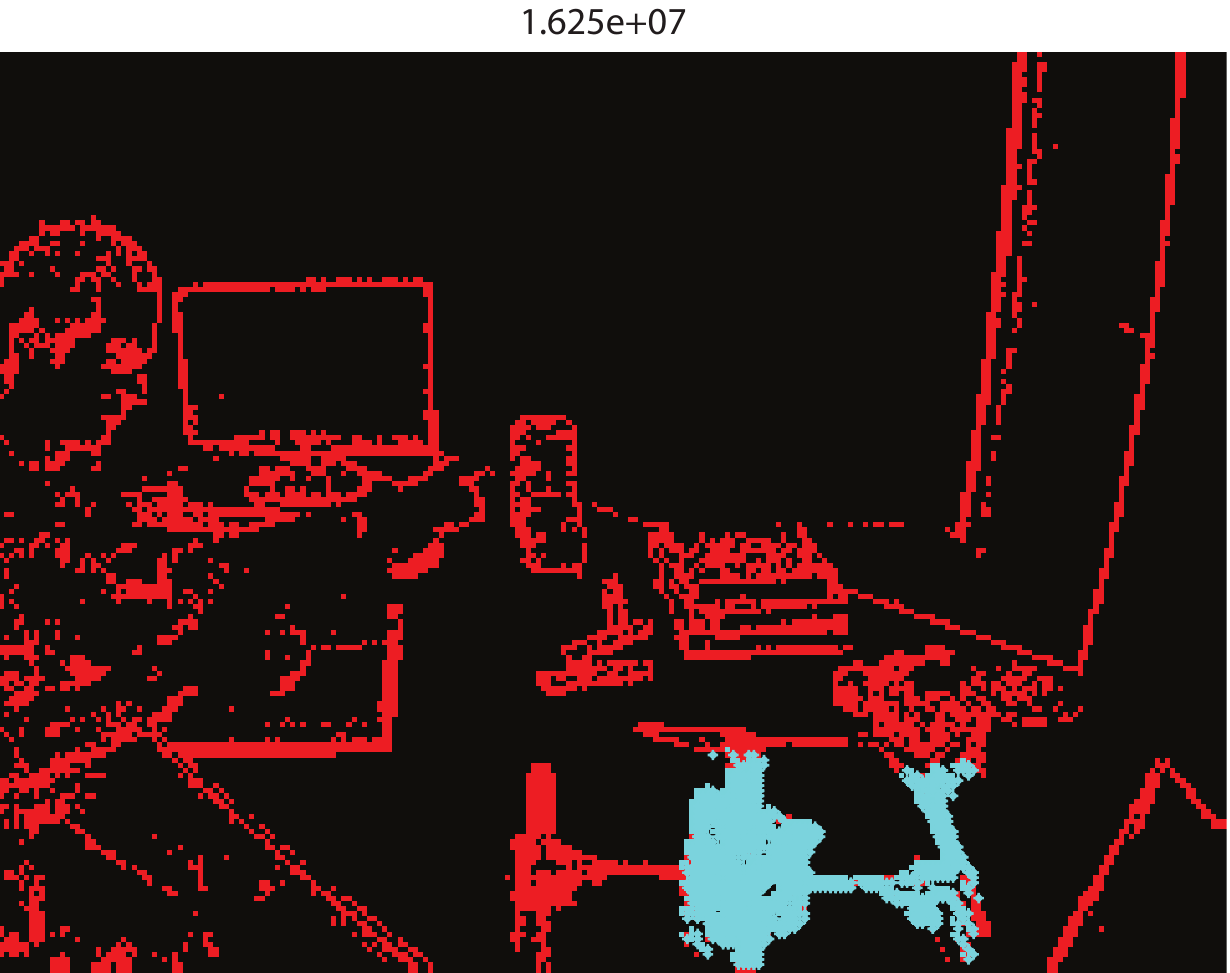}%
\label{fig5_fourth_case}}
\hfil
\subfloat[Tracked:10.11\% Total:11790]{\includegraphics[width=3.5cm]{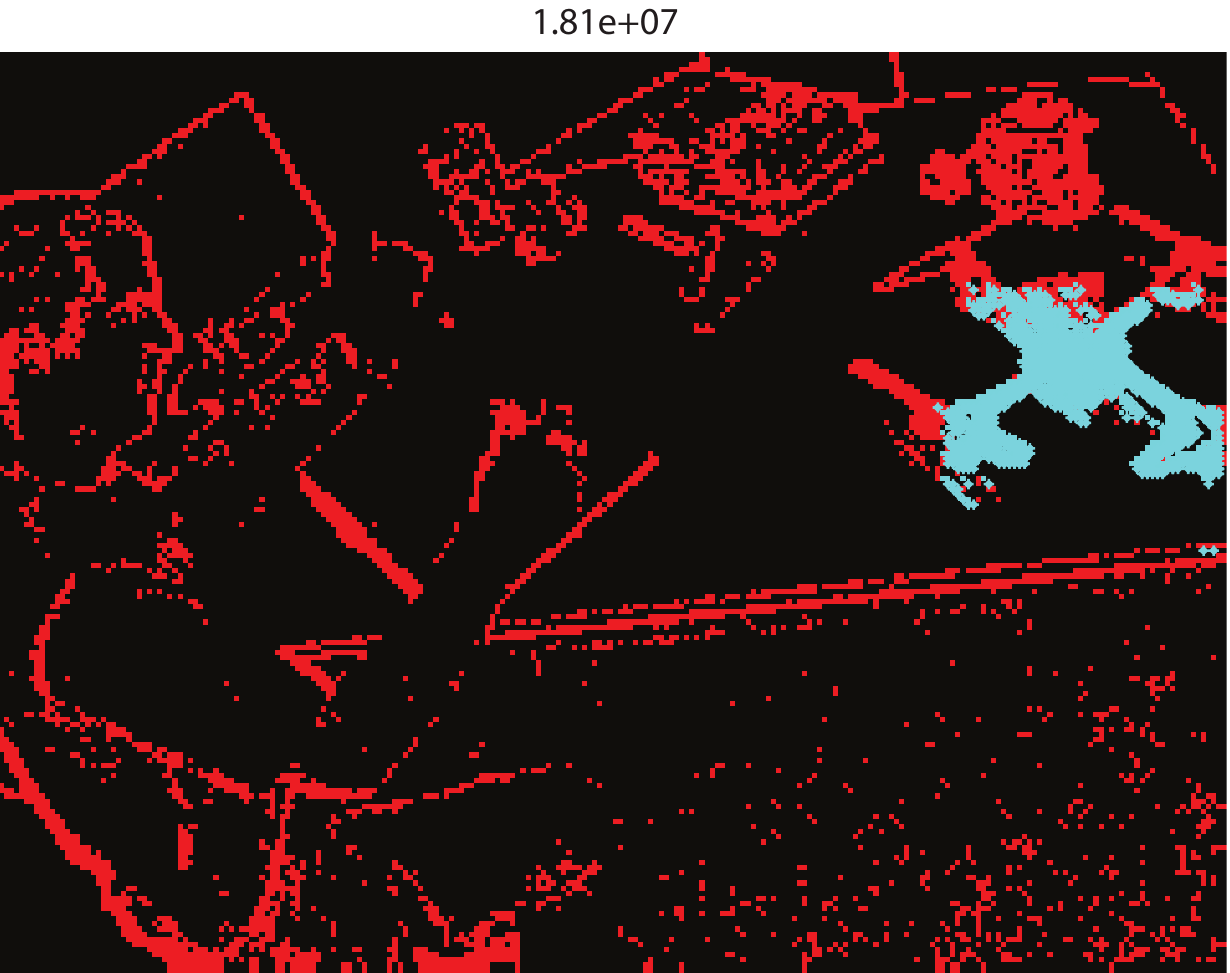}%
\label{fig5_fifth_case}}
\caption{Tracking of the drone object under general 6-DOF camera motion.}
\label{fig:6doftrackingresults}
\endgroup
\end{figure*}
\section{Experiments}
\label{sec:results}
For testing the proposed {\em e-TLD} system, the dynamically captured data in the event-camera dataset \cite{Mueggler2017} was used. For each object, the training ROI was manually specified during the first 500ms of the recording and the testing was done until the end of the recording (60s). Using the ground truth annotations we created, it is possible to quantitatively evaluate the tracking performance and this sets up one of the first realistic tracking benchmarks for the neuromorphic vision community. The object location is specified as a bounding box within a short time-interval of 10ms for the full-length of the data. The ground truth annotations for quantitative performance evaluation are available online\footnote{\url{https://github.com/nusneuromorphic/Object_Annotations}}. 
\par
In general, tracking algorithms are evaluated by two metrics \cite{wu2013online}, which are center location error (CLE) and overlap success (OS). The first metric, CLE, indicates the average Euclidean distance between the ground-truth and the estimated center location (in $pixels$). The second metric, OS, is defined as the number of times (\%) the tracker output overlaps with the ground truth annotations while having a minimal overlap of 50\%. We use OS as the primary metric for our evaluation and we report the results at a threshold of 50\%, which correspond to the PASCAL evaluation criteria. In addition, we also report CLE when there is an overlap success to show the closeness of ground truth match. 
\subsection{Parameters}
For each object, an ROI was manually specified during the first 500ms of the recording (training data) and the rest of the recording was used for testing. A codebook size of $K=500$ is used to build the object and the background representation. For the local search tracker, an important parameter is the tracker confidence $\tau_t \in [0,1]$, which is typically set to a value close to the average tracking score required for successful track. Nonetheless, we report the tracking performance for various thresholds in the range $[0.5, 1]$. The SVM training is performed with Bayesian bootstrapping that outputs equal number of samples as the initial number of descriptors. For example, if there are $N_1 = 840$ ROI descriptors at the user initialization state, $N_1$ samples having $N_1/2$ descriptors in each sample are obtained after bootstrapping. The parameter $\tau$ of the detector is the same as the tracker threshold, although the window size is the whole image plane instead of the tracked region. The system performance is also reported by varying $\tau$  in steps.

\subsection{Results on Event Camera Dataset}
\label{subsec:results}
As shown in Figure \ref{fig:trackingresults}, {\em e-TLD} is able to track and detect objects of various sizes and shapes. In these results, the overlaid markers indicate the position of the tracked object in the field-of-view of the event camera. Although the appearance of the object changed considerably during the translational camera motion, rotation was intentionally kept minimal in these recordings. Separate recordings of the same scene are available for the general 6-DOF camera motion, which induces drastic view-point change of the object. Figure \ref{fig:6doftrackingresults} shows the tracking of the drone object under drastic view-point variations, showing robustness of the {\em e-TLD} system also to induced scale and rotation changes. For a qualitative viewing of the results, the web video\footnote{{\em e-TLD} demo (updated): \url{https://youtu.be/kkw69aVOoJY}} clearly shows the fine grain monitoring of the object as long as it is in the FOV where rotation and scale changes can be monitored progressively because of the online SVM learning and micro-second sensor resolution. 
\par
As seen from the video results, although the local sliding process is dependent on the event activity rather than a time-based tracking process, the tracker loses the object during the initial stages when abrupt changes in location and appearance happen due to rotation and viewpoint changes. Nonetheless, towards the end of each recording, especially for the rotation and 6-DOF motion, the tracker becomes tolerant as the object under different variations has been included by the online learning step. Also worth noting is since we are processing a fixed number of events, as a percentage of events inside the tracking window or the global field-of-view for the detector, the faster speed of motion as the recording progresses only results in faster processing, and does not affect functionality. Note that the static drone object has a good performance for all three motion profiles compared to the rest of the objects, being closer to the camera and larger in relative size.
\par
Table.~\ref{tab:dyntrack} shows the performance of {\em e-TLD} using the publicly available event-camera dataset \cite{Mueggler2017}. The translation motion profile results in the best average OS compared to rotation and 6-DOF motion. This is partly due to the local sliding tracker update that is looking for a rectangle within a search space (Fig.~\ref{fig:locslide}), so naturally the translation motion entails that objects can be fully captured within the candidate bounding boxes. However, it is interesting to note that the average center location error is highest for the translation case because of incomplete overlap with the actual object. On the other hand, there are no clear indications on how the size of the object affects CLE even though tracking and detecting smaller objects, such as the cup, was difficult. In fact, the OS score for the cup is lowest among the objects, especially for the 6-DOF case.  
\par
In addition to the tougher rotation and 6-DOF motion profile, the dynamic human moves farther away from the camera in the recordings after taking the cup in his hand. This induces tracker fail for a longer period of time and also unable to precisely detect the tracked object (lower OS). It is worth stating that the low-resolution of the DAVIS240C further reduces the event density for such objects and makes detecting farther away objects challenging. Neuromorphic vision sensors with higher resolution \cite{Posch2008} could alleviate these issues to a considerable degree. 

\begin{table}[t]
\begin{center}
\caption{Quantitative tracking results of {\em e-TLD}.}
\label{tab:dyntrack}
\begin{tabular}{ccccccc}
\hline\noalign{\smallskip}
Trans. &  Head & Monitor & Drone & Cup & Books & \textit{Avg.}\\
\noalign{\smallskip}
\hline
\noalign{\smallskip}
{\bf OS}  &	0.5895 &	0.7643 &	0.8731  &	0.5972 &	0.6809 & \textit{$0.7010$}  \\
{\bf CLE}  &	2.4437  &	2.2515 &	1.8430  &	0.6773 &	0.9352 & \textit{$1.6301$}  \\
\hline
\\
\end{tabular}

\begin{tabular}{ccccccc}
\hline\noalign{\smallskip}
Rot. &  Head & Monitor & Drone & Cup & Books & \textit{Avg.} \\
\noalign{\smallskip}
\hline
\noalign{\smallskip}
{\bf OS}  &	0.1832  &	0.4878 &	0.5331  &	0.4777 &	0.5083  & \textit{$0.4380$} \\
{\bf CLE}  &	0.6414  &	2.8829 &	1.8000  &	0.6492 &	1.3332  & \textit{$1.4613$} \\
\hline
\\
\end{tabular}

\begin{tabular}{ccccccc}
\hline\noalign{\smallskip}
6-DOF &  Head & Monitor & Drone & Cup & Books & \textit{Avg.} \\
\noalign{\smallskip}
\hline
\noalign{\smallskip}
{\bf OS}  &	0.3172  &	0.6174 &	0.5110  &	0.2247 &	0.2933  & \textit{$0.3927$}  \\
{\bf CLE}  &	0.9879  &	2.4385 &	2.0231  &	0.3204 &	1.1778  & \textit{$1.3895$} \\
\hline
\\
\end{tabular}
\end{center}
\end{table}
\par
Finally, using the same parameters for the tracker representation and thresholds, \cite{Ramesh2018} obtains a mean overlap score of $0.5841$ for the objects (`Head' $0.4591$, `Monitor' $0.7645$, `Drone' $0.8217$, `Cup' $0.2390$ and `Books' $0.6364$) compared to mean OS of $0.7010$ in Table.~\ref{tab:dyntrack} using the dynamic translation data. In other words, the tracking has been improved by 10\% compared to our previous framework. This has been made possible by parameter sharing between the tracker and detector as outlined in Sec.~\ref{subsec:detectionphase}. 

\begin{table}[t]
\begin{center}
\caption{Comparison to e-LOT \cite{Ramesh2019} tracking using the OS metric.}
\label{tab:dyntrack2}
\begin{tabular}{ccccccc}
\hline\noalign{\smallskip}
Trans. &  Head & Monitor & Drone & Cup & Books & \textit{Avg.}\\
\noalign{\smallskip}
\hline
\noalign{\smallskip}
{\bf e-LOT }  &	0.1450  &	0.1796 &	0.2021  &	\textbf{0.7400} &	0.2647 & 0.3063  \\
{\bf e-TLD}  &	\textbf{0.5895} &	\textbf{0.7643} &	\textbf{0.8731}  &	0.5972 &	\textbf{0.6809} & \textbf{0.7010}  \\
\hline
\\
\end{tabular}

\begin{tabular}{ccccccc}
\hline\noalign{\smallskip}
Rot. &  Head & Monitor & Drone & Cup & Books & \textit{Avg.} \\
\noalign{\smallskip}
\hline
\noalign{\smallskip}
{\bf e-LOT }  &	0.1349  &	0.1503 &	0.2445  &	0.4359 &	0.4115  & 0.2754 \\
{\bf e-TLD}  &	\textbf{0.1832}  &	\textbf{0.4878} &	\textbf{0.5331}  &	\textbf{0.4777} &	\textbf{0.5083}  & \textbf{0.4380} \\
\hline
\\
\end{tabular}

\begin{tabular}{ccccccc}
\hline\noalign{\smallskip}
6-DOF &  Head & Monitor & Drone & Cup & Books & \textit{Avg.} \\
\noalign{\smallskip}
\hline
\noalign{\smallskip}
{\bf e-LOT }  &	0.0612  &	0.2384 &	0.0244  &	\textbf{0.3156} &	0.0692  & 0.1417 \\
{\bf e-TLD}  &	\textbf{0.3172}  &	\textbf{0.6174} & \textbf{0.5110}  &	0.2247 &	\textbf{0.2933}  & \textbf{0.3927}  \\
\hline
\\
\end{tabular}
\end{center}
\end{table}
\subsection{Comparison to state-of-the-art}
\label{sec:elotcompare}
The descriptor proposed in \cite{Ramesh2019} for event cameras was demonstrated with promising results for four different vision problems, namely object classification, tracking, detection and feature matching, as also noted in \cite{Linares-Barranco2019}. However, the event-based long-term object tracking (e-LOT)\cite{Ramesh2019} solution assumes a clean background surrounding the objects for tracking with a less reliable detection approach. Nonetheless, e-LOT is the only comprehensive work for event cameras addressing the problem of long-term object tracking and thus we make a comparison to e-TLD. 
\par
Table.~\ref{tab:dyntrack2} compares e-LOT with the general purpose e-TLD framework for dynamic object tracking on the event camera dataset using the main OS metric. It is clear that e-LOT does not generalize well to cluttered and more generic data. In all three motion cases, e-TLD comprehensively outperforms e-LOT using the average OS score while performing slightly underpar for the `cup' object. We attribute this anomaly to the tailor-made e-LOT system for tracking objects with cleaner background, which the `cup' object encounters due to its unique placement in the scene compared to the other objects. 

\subsection{Real-time testing}
\label{subsec:realtime}
The {\em e-TLD} framework was implemented in C++ using Visual Studio IDE for Windows 10 with several practical design considerations. For instance, the feature descriptor encodes the distribution of the events using a fixed log-polar lattice \cite{Ramesh2017a}. Thus, instead of generating the log-polar grid at every new event position, the event is transposed to a known location (say top left of the image), and the features are obtained using a pre-computed log-polar grid to save computational time. In addition, the local sliding window operation is efficiently accomplished by maintaining distinct tracker representations for each candidate window. Then, using a look-up table that is computed offline for ascertaining whether an event belongs to a rectangular candidate window, the respective tracker representations are updated. After an instance of successful track, the tracker representations are reset and updated according to incoming events. Similarly for the detector, the global sliding window can be implemented by counting the detected events using a look-up table as they arrive instead of waiting to detect after the threshold $\tau$ is reached. 
\par
Real-time testing was carried out using a DAVIS camera, interfaced and powered by a workstation running an Intel Core$\texttrademark$ i7 3.6GHz processor. Our implementation running on a single thread achieves an average latency of $45\mu s$ per incoming event, which is about $140\times$ faster than \cite{tracking3_movingobjIROS2018}. A standard global shutter camera is likely to generate images with motion blur artifacts while the handheld camera is constantly in motion. However, the event camera and our algorithm are able to track the object, as shown in this video\footnote{Real-time demo (updated): \url{https://tinyurl.com/ske6nk7}}. 
\par
There are three parts to the above demo video. Firstly, handheld testing of the detection system (without tracking) was done to showcase real-time performance with the computationally more intensive global sliding window step. This is to highlight the possibility of a tracking-by-detection approach, and mainly to showcase that even in extreme cases where local sliding window is inhibited, the global sliding window detector can still output object location in real-time. Subsequently, we show that e-TLD still works for simple shapes data, followed by real-time performance on the dynamic data. We would like to point out that for the shapes and dynamic dataset recordings, they were input to the C++ system by simulating the camera interface, and thereby enabling real-time output. 

\begin{figure}[t]%
\centering
\includegraphics[width=0.9\columnwidth]{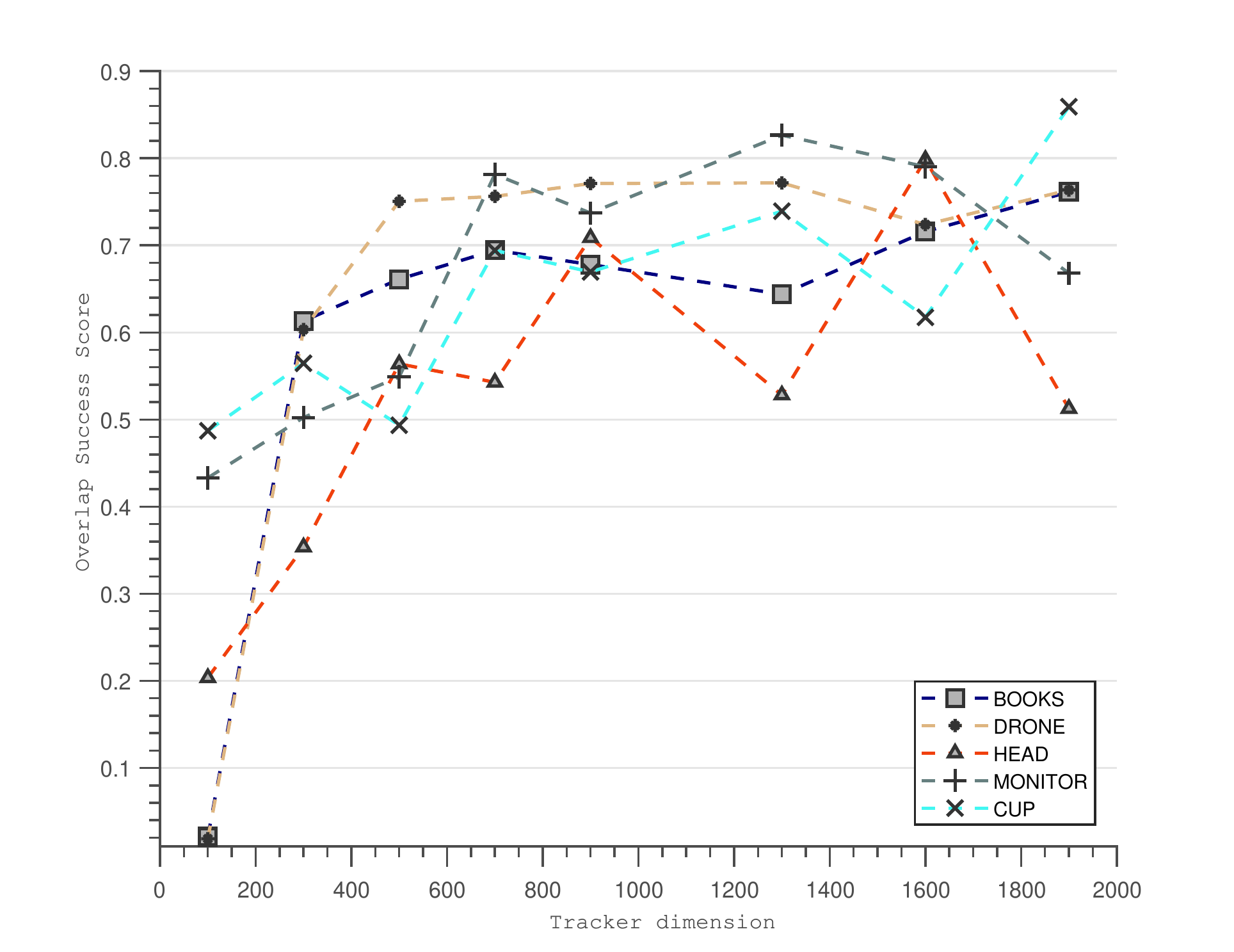}%
\caption{Varying the dimension of the object representation.}%
\label{fig:code_vs_os}%
\end{figure}

\begin{figure}[t]%
\centering
\includegraphics[width=0.9\columnwidth]{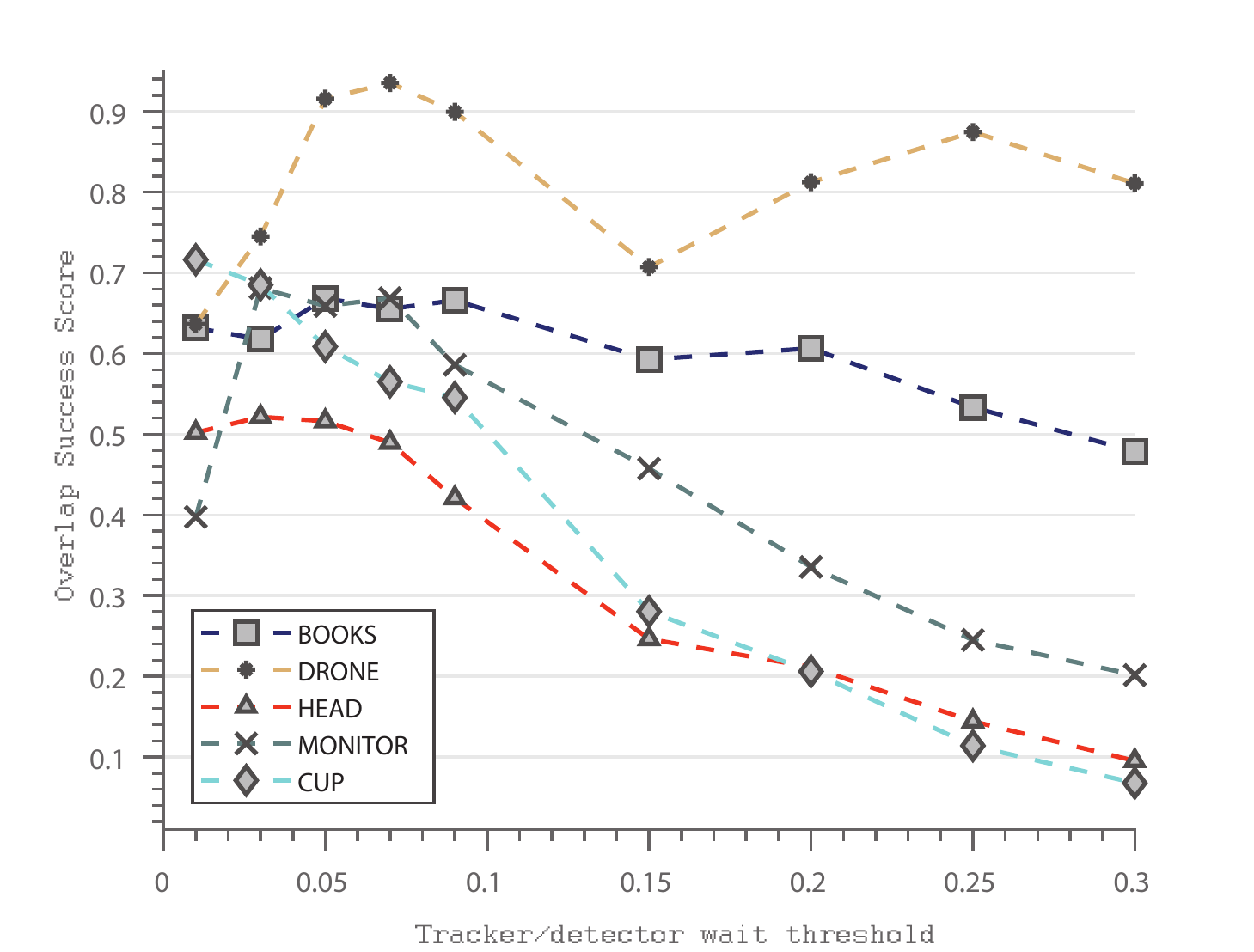}%
\caption{Varying tracker/detector wait threshold $\tau$.}%
\label{fig:tau_vs_os}%
\end{figure}

\begin{figure}[t]%
\centering
\includegraphics[width=0.9\columnwidth]{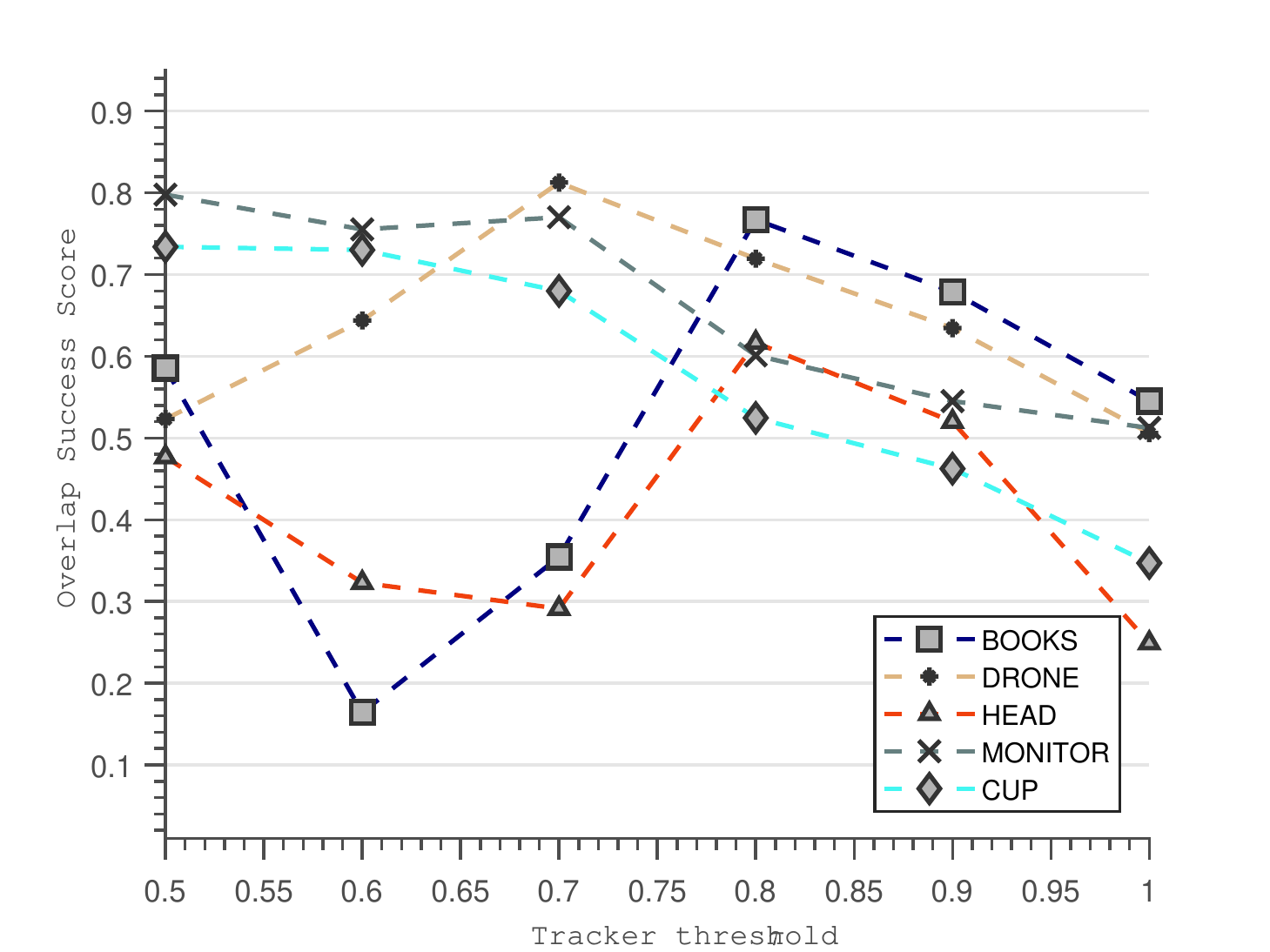}%
\caption{Varying tracker threshold $\tau_t$.}%
\label{fig:tauthresh_vs_os}%
\end{figure}

\begin{figure}[t]%
\centering
\includegraphics[width=0.9\columnwidth]{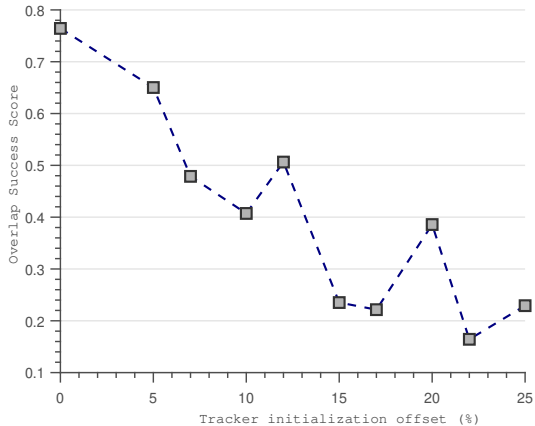}%
\caption{Sensitivity of {\em e-TLD} to initialization location for the monitor object.}%
\label{fig:noise_vs_os}%
\end{figure}
\section{Discussion}
\label{sec:discuss}
In this section, we report the {\em e-TLD} system performance by varying crucial algorithm parameters and draw insights for setting them for different object and motion scenarios. The translation data is used to uniformly study the system performance in the experiments reported below. 
\par
The dimension of the object representation has a direct impact on the tracker performance (eq. \eqref{eq:newBOW}). In general, a higher dimensional discriminative representation is expected to provide better background vs. object separation, as noted for the object classification task in \cite{Ramesh2017a}. Fig.~\ref{fig:code_vs_os} confirms the trend with increase in overlap success for higher dimensions. In particular, objects such as the books, drone and cup exhibit further increase beyond 1500 dimensions although saturation is expected. It is worth noting that all the reported results in the previous section was with 500 visual words as a compromise between tracker performance and running time. Our intention is to foster new research in this niche domain instead of reporting the best OS values on the dynamic scenes of the event camera dataset. 
\par
Another crucial parameter is the ``waiting time" of the tracker and detector, $\tau$, set as a percentage of the events received within the tracker state or the whole image plane in the case of the detector. Fig.~\ref{fig:tau_vs_os} shows a steady drop in the overlap success for $\tau$ beyond $0.1$ up to $0.3$ except for the drone object. This parameter was set to $0.05$ universally for the reported experiments in Sec.~\ref{subsec:results}. The detector plays a crucial role in re-capturing the object and in some cases we found that objects uniform in intensity, though large in size, like the computer monitor, could only be partially detected on most occasion due to the corresponding low event generation. However, objects that are not ``hollow" like the drone can tolerate even high values of $\tau$ as the density of events is high. 
\par
Next, the tracker threshold $\tau_t$ that determines track success or failure is varied in Fig.~\ref{fig:tauthresh_vs_os}. It is expected that a value very close to the average tracker score, $\tau_t > 0.9$, is very strict compared to allowing background objects to be tolerated by setting $\tau_t < 0.6$. Therefore, $\tau_t \in [0.6, 0.8]$ is practical in many scenarios where a smooth track is expected rather than frequent switching to the detector. In the experiments in Sec.~\ref{subsec:results}, a value of $0.8$ was used to report the OS and CLE measures. Objects like the head and drink cup benefit from having low thresholds as detecting them is difficult compared to static objects like the drone. 
\par
Finally, it is of interest to examine the robustness of the {\em e-TLD} system to object initialization offset, which is a possible scenario in practical applications. In other words, since relative motion from object boundaries would generate a lot of events, if an initialization is imperfect and does not perfectly capture the entire object extent, the tracker may not perform as expected. Figure.~\ref{fig:noise_vs_os} shows a gradual decreasing performance of the {\em e-TLD} system up to 25\% offset from the ground truth initial location for the monitor object in the translation setting. From this experiment, it is safe to conclude that a reasonable system performance can be expected for up to 10\% drift in bounding box coordinates during system initialization. 
\par
While the object representation needs to be learnt within a reasonable 10\% drift limit for system initialization, it is not imperative that the initial object location has to be labeled by a user. As the neuromorphic vision community matures in developing generic object detectors for real-world objects, detecting instances of commonly seen objects without prior annotation will be feasible. In which case, {\em e-TLD} will be initialized by a detector, internal or external to the framework. In other words, the e-TLD framework is not inherently limited by the need for user intervention for initialization. Similarly, the binary classification scheme is also less of a limitation in the long haul when we can develop capabilities and implementations that can simultaneously run multiple trackers, specific to each object, in hardware implementations (e.g. \cite{ussa2020hybrid} runs up to eight trackers concurrently) or in pure software with sufficient parallel processing power.

\section{Conclusion}
\label{sec:conc}
This paper presented a long-term object tracking system for event cameras, showing how an event-based tracker and detector permits the application of an event camera to the important problem of long-term object tracking, and hopefully this opens the door to similar approaches for other related vision problems. The tracker uses an event-based {\em local sliding window} technique that performs reliably in scenes with cluttered and textured background. In addition, Bayesian bootstrapping is used to assist real-time processing and boost the discriminative power of the object representation. On the other hand, when the object re-enters the field-of-view of the camera, a \emph{data-driven, global sliding window} based detector locates the object under different view-point conditions for subsequent tracking. Extensive experiments on a publicly available event camera dataset demonstrates the ability to track and detect arbitrary objects of different shapes and sizes under various motion profiles. Using the ground truth locations we created, quantitative measurement is reported for the event-based tracking method with critical insights on various performance issues. Finally, we showcase the real-time object tracking performance of e-TLD using a C++ implementation for scale, rotation, view-point and occlusion scenarios in a lab setting. It  is  worth  restating  that  the  data  rate  of  the  DAVIS  event
camera  used  in  our  experiments  is  typically  in  the  order  of 150  KB/s  while  a  standard  grayscale  VGA  camera  outputs
frames  at  30Hz  or  about  10MB/s.  The  only  information  that
is  important  for  tracking  and  detection  is  how  edges  move,
and the event camera naturally outputs this information while
sidestepping problems of blur, low-dynamic range and limited
motion information that standard cameras create.

\ifCLASSOPTIONcaptionsoff
  \newpage
\fi




\bibliographystyle{IEEEtran}
\bibliography{images/mybib}

\begin{IEEEbiography}[{\includegraphics[width=1in,height=1.25in,clip,keepaspectratio]{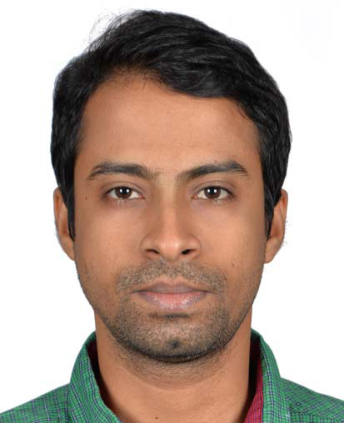}}]{Bharath Ramesh}
 received the B.E. degree in electrical \& electronics engineering from Anna University of India in 2009; M.Sc. and Ph.D. degrees in electrical engineering from National University of Singapore in 2011 and 2015 respectively, working at the Control and Simulation Laboratory on Image Classification using Invariant Features. Bharath’s main research interests include pattern recognition and computer vision. At present, his research is centered on event-based cameras for autonomous robot navigation.
\end{IEEEbiography}

\begin{IEEEbiography}[{\includegraphics[width=1in,height=1.25in,clip,keepaspectratio]{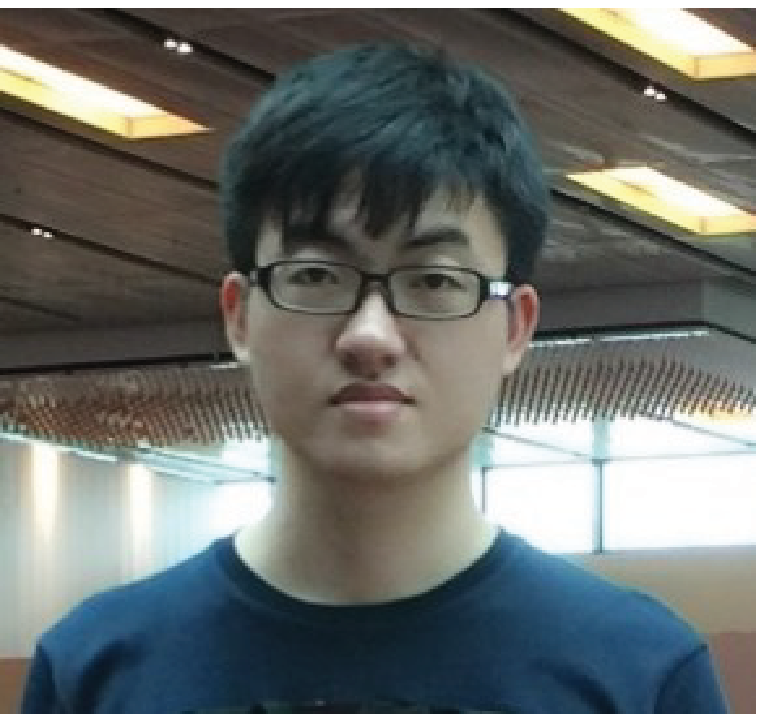}}]{Shihao Zhang}
 is currently an undergraduate at National University of Singapore, studying under the double degree program of computer engineering and economics. He is also under a research intern in Temasek Lab, with focus on event-based object tracking and dealing with real-time problems concerning event-based visual odometry.
\end{IEEEbiography}

\begin{IEEEbiography}[{\includegraphics[width=1in,height=1.25in,clip,keepaspectratio]{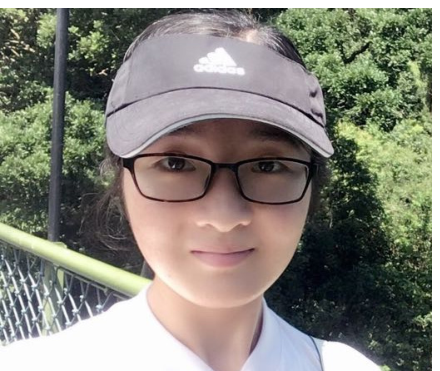}}]{Hong Yang}
 received her Bachelor's degree at University of Electronic Science and Technology of China (UESTC). She was a master student of NUS and under a working scheme in Temasek Lab to perform research on event-based cameras, dealing with real-time pattern recognition problems.
\end{IEEEbiography}

\begin{IEEEbiography}[{\includegraphics[width=1in,height=1.25in,clip,keepaspectratio]{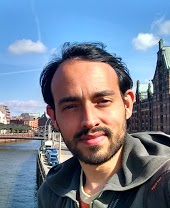}}]{Andres Ussa}
received his MSc from TU Kaiserslautern and University of Southampton in Embedded Computing Systems. His research experience has been focused on embedded systems design and machine learning applications. He had a short experience as a Software/Hardware Developer for consumer electronics.
\end{IEEEbiography}

\begin{IEEEbiography}[{\includegraphics[width=1in,height=1.25in,clip,keepaspectratio]{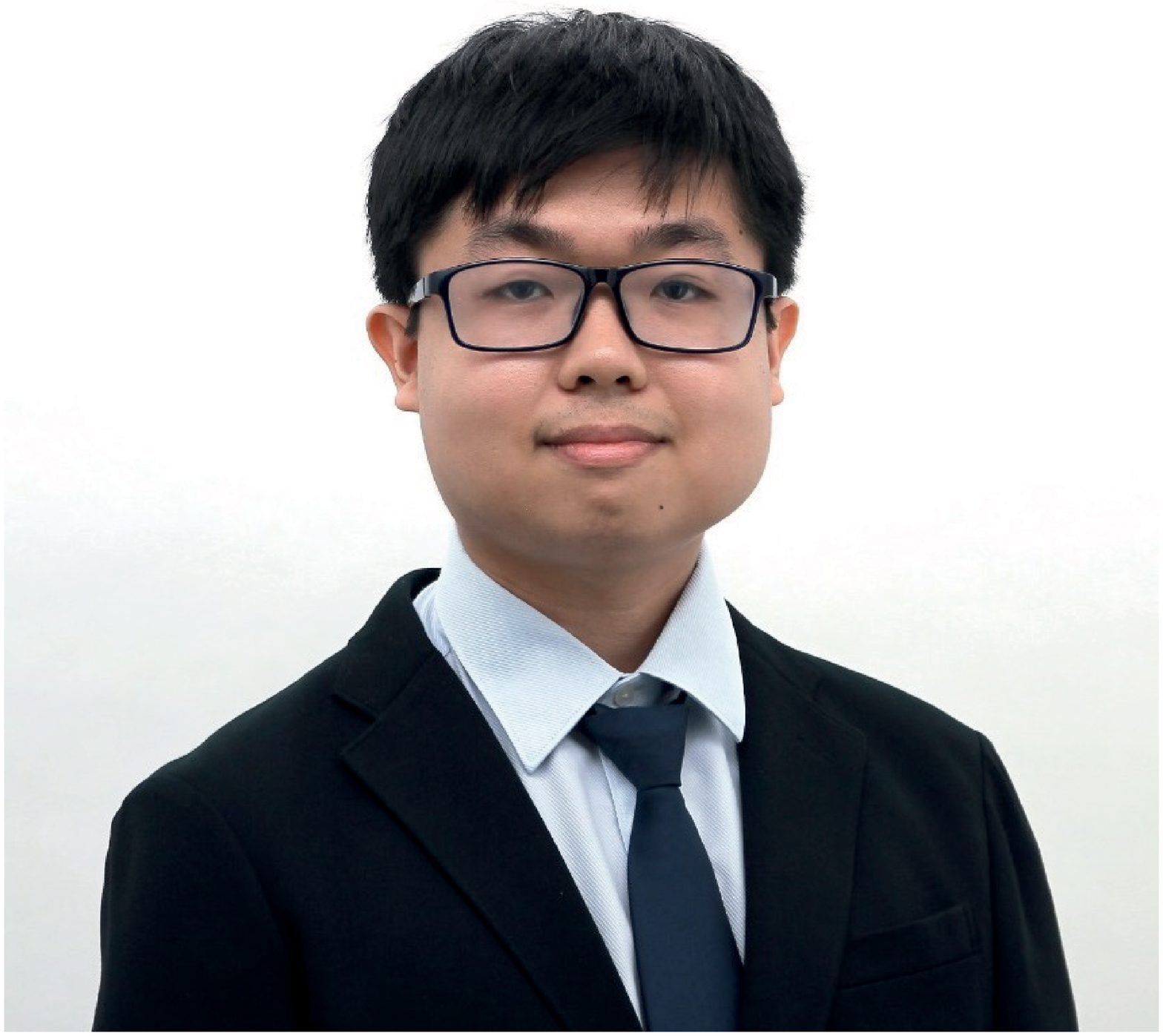}}]{Matthew Ong}
 is currently an undergraduate at National University of Singapore, studying in the department of electrical and computer engineering. His final year project focuses on event-based vision, dealing with tracking issues under dynamic camera motion profile.
\end{IEEEbiography}

\begin{IEEEbiography}[{\includegraphics[width=1in,height=1.25in,clip,keepaspectratio]{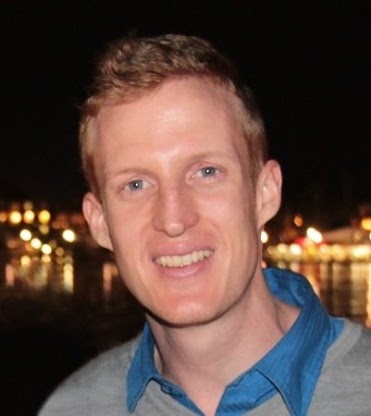}}]{Garrick Orchard}
 holds a B.Sc. degree (with honors, 2006) in electrical engineering from the University of Cape Town, South Africa and M.S.E. (2009) and Ph.D. (2012) degrees in electrical and computer engineering from Johns Hopkins University, Baltimore, USA. His research focuses on developing neuromorphic vision algorithms and systems for real-time sensing on mobile platforms. His other research interests include mixed-signal very large scale integration (VLSI) design, compressive sensing, spiking neural networks, visual perception, and legged locomotion.
\end{IEEEbiography}

\begin{IEEEbiography}[{\includegraphics[width=1in,height=1.25in,clip,keepaspectratio]{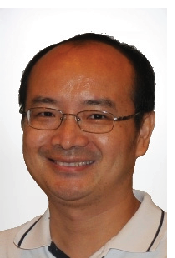}}]{Cheng Xiang}
 received the B.S. degree in mechanical engineering from Fudan University, China in 1991; M.S. degree in mechanical engineering from the Institute of Mechanics, Chinese Academy of Sciences in 1994; and M.S. and Ph.D. degrees in electrical engineering from Yale University in 1995 and 2000, respectively. He is an Associate Professor in the Department of Electrical and Computer Engineering at the National University of Singapore. His research interests include computational intelligence, adaptive systems and pattern recognition.
\end{IEEEbiography}
\vfill




\end{document}